%% file: anonymous-submission-latex-2025.tex
\documentclass[letterpaper]{article} % DO NOT CHANGE THIS
\usepackage{aaai25}  % DO NOT CHANGE THIS
\usepackage{times}  % DO NOT CHANGE THIS
\usepackage{helvet}  % DO NOT CHANGE THIS
\usepackage{courier}  % DO NOT CHANGE THIS
\usepackage[hyphens]{url}  % DO NOT CHANGE THIS
\usepackage{graphicx} % DO NOT CHANGE THIS
\usepackage{natbib}  % DO NOT CHANGE THIS AND DO NOT ADD ANY OPTIONS TO IT
\usepackage{caption} % DO NOT CHANGE THIS AND DO NOT ADD ANY OPTIONS TO IT
\usepackage{algorithm}
\usepackage{algorithmic}

\usepackage[dvipsnames, svgnames, x11names]{xcolor}
\usepackage{booktabs}
\usepackage{multirow} 
\usepackage{colortbl}
\newcommand{\gray}[1]{\textcolor{gray}{#1}}
\newcommand{\graysmall}[1]{\textcolor{gray}{\small#1}}
\usepackage{pifont}
\definecolor{my_green}{rgb}{0.252, 0.736, 0.240}
\definecolor{my_red}{rgb}{0.950, 0.250, 0.250}
\newcommand{\gcmark}{\color{my_green}{\ding{51}}} % Green check mark
\newcommand{\rxmark}{\color{my_red}{\ding{55}}} % Red cross mark

\usepackage{arydshln}
\newcommand{\hrulespace}{\rule{0pt}{1em}}
\newcommand{\largehrulespace}{\rule{0pt}{1.1em}}
\newcommand{\tstart}[1]{\textcolor{gray}{\footnotesize\texttt{<\kern-0.2ptT\scalebox{0.85}{#1}\kern-0.3pt>}}}
\newcommand{\tend}[1]{\textcolor{gray}{\footnotesize\texttt{<\kern-1pt/\kern-0.6ptT\scalebox{0.85}{#1}\kern-0.3pt>}}}
\definecolor{my_orange}{RGB}{215,138,24}
\newcommand{\gtokens}{\textcolor{my_orange}{\small\texttt{\,<T>\,</T>}}}
\newcommand{\thinhline}{\noalign{\hrule height 0.1pt}}

\newcommand{\action}[1]{\textcolor[HTML]{0074d9}{#1}}
\newcommand{\object}[1]{\textcolor[HTML]{3d9970}{#1}}
\newcommand{\location}[1]{\textcolor[HTML]{cd7c0c}{#1}}

\newcommand{\bigdatasetname}{\textsc{\Large M\small ULTI\Large H\small OP\Large -E\small GO\Large QA}}
\newcommand{\datasetname}{\textsc{\normalsize M\footnotesize ULTI\normalsize H\footnotesize OP\normalsize -E\footnotesize GO\normalsize QA}}
\newcommand{\tabledatasetname}{\textsc{\normalsize M\scriptsize ULTI\normalsize H\scriptsize OP\normalsize -E\scriptsize GO\normalsize QA}}
\newcommand{\smalldatasetname}{\textsc{\small M\scriptsize ULTI\small H\scriptsize OP\small -E\scriptsize GO\small QA}}
\newcommand{\modelname}{\textbf{GeLM}}

\newcommand{\eg}{{\em e.g.}}
\newcommand{\ie}{{\em i.e.}}
\usepackage{amsmath}
\usepackage{amssymb}
\newcommand{\garrow}[1]{\,\textcolor[HTML]{247634}{\scriptsize\textbf{$\uparrow$}\,{\bf\scriptsize #1}}}

\usepackage{tcolorbox}
\usepackage{varwidth}
\newcommand{\VarSty}[1]{\textnormal{\ttfamily\color{blue!90!black}#1}\unskip}
\usepackage{subcaption}

\usepackage{mdframed}

\usepackage{hyperref} % need to be removed with comments in .sty
\definecolor{darkblue}{rgb}{0, 0, 0.5}
\hypersetup{colorlinks=true, citecolor=darkblue, linkcolor=darkblue, urlcolor=darkblue}
\nocopyright % need to be removed
\usepackage{newfloat}
\usepackage{listings}
\DeclareCaptionStyle{ruled}{labelfont=normalfont,labelsep=colon,strut=off} % DO NOT CHANGE THIS
\floatstyle{ruled}
\newfloat{listing}{tb}{lst}{}
\floatname{listing}{Listing}
\title{Grounded Multi-Hop VideoQA in Long-Form Egocentric Videos}
\author{
    Qirui Chen, Shangzhe Di, Weidi Xie
}
\affiliations{
    Shanghai Jiao Tong University 
    \\[3pt]
    \{chen\_qirui, dishangzhe, weidi\}@sjtu.edu.cn
    \\[.5em]
    \url{https://qirui-chen.github.io/MultiHop-EgoQA}
}

\begin{document}

\maketitle
\input{sections/0_abstract}
\input{sections/1_introduction}
\input{sections/2_related_work}
\input{sections/3_curation}

\input{sections/4_architecture}

\input{sections/5_experiments}
\input{sections/6_conclusion}
% \clearpage

% \input{sections/checklist}

\input{sections/7_appendix}

\vspace{20pt}
\bibliography{aaai25}

\end{document}

%% file: sections/0_abstract.tex
\begin{abstract}

This paper considers the problem of \emph{Multi-Hop Video Question Answering (MH-VidQA)} in long-form egocentric videos. 
This task not only requires to answer visual questions, 
but also to localize multiple relevant time intervals within the video as visual evidences. We develop an automated pipeline to create multi-hop question-answering pairs with associated temporal evidence, 
enabling to construct a large-scale dataset for instruction-tuning. 
To monitor the progress of this new task, we further curate a high-quality benchmark, \textbf{\smalldatasetname}, with careful manual verification and refinement. Experimental results reveal that existing multi-modal systems exhibit inadequate multi-hop grounding and reasoning abilities, resulting in unsatisfactory performance. 
We then propose a novel architecture, 
termed as \textbf{G}rounding Scattered \textbf{E}vidence with Large \textbf{L}anguage \textbf{M}odel~(\modelname), that enhances multi-modal large language models (MLLMs) by incorporating a grounding module to retrieve temporal evidence from videos using flexible grounding tokens. 
Trained on our visual instruction-tuning data, \modelname~demonstrates improved multi-hop grounding and reasoning capabilities, setting a new baseline for this challenging task. Furthermore, when trained on third-person view videos, the same architecture also achieves state-of-the-art performance on the single-hop VidQA benchmark, ActivityNet-RTL, demonstrating its effectiveness.
\end{abstract}


%% file: sections/1_introduction.tex
\section{Introduction}

With the rapid development of computer vision, the community has witnessed a significant interest in deploying vision systems within embodied agents, such as autonomous vehicles and humanoid robots. 
In such scenarios, the inputs are typically long, continuous video streams from a first-person perspective, capturing the world through the eyes of an agent actively interacting with its environment. 
For the virtual assistants or physical robots to be useful, the ability to perform egocentric video question answering~(VidQA) is crucial, stemming from two aspects: 
{\em first}, VidQA leverages language as a natural interface for human-machine interaction, thereby enhancing the usability and accessibility for the general public; 
{\em second}, it can encompass various vision tasks about the `who', `when', `where', and `what' of an individual's daily life, 
{\em e.g.}, action recognition, object detection, and scene understanding, thus acting as a robust and comprehensive benchmark for video understanding.

\begin{figure}[t]
    \centering
    \includegraphics[width=\linewidth]{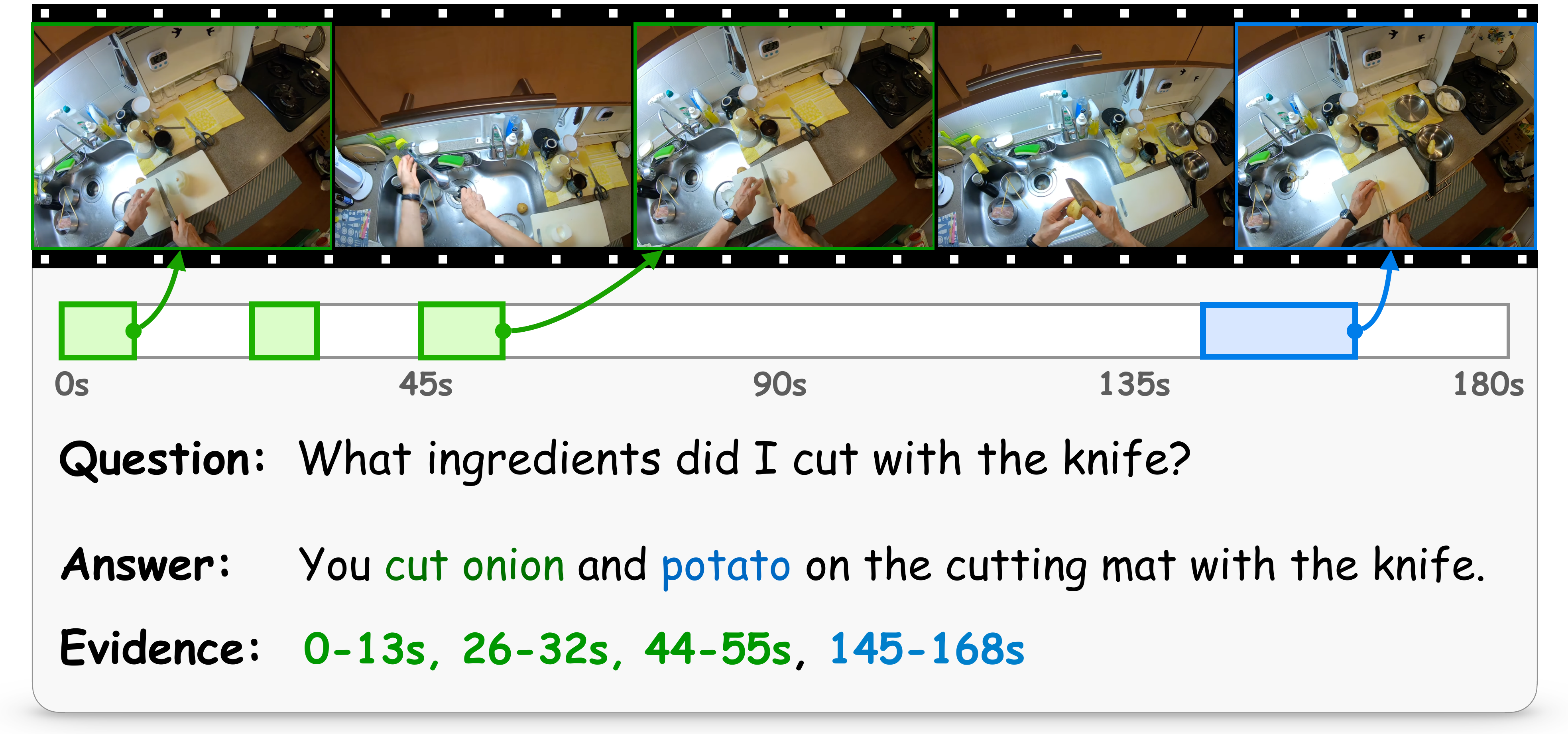}
    \vspace{-0.55cm}
    \caption{We introduce the problem of \textit{Multi-Hop Video Question Answering} for long-form egocentric video understanding. This task requires the model to answer questions by gathering and reasoning across scattered visual clues, necessitating the grounding of multiple relevant time spans as supporting evidence.}
    \label{fig:teaser}
    \vspace{-0.5cm}
\end{figure}

Recently, the introduction of Ego4D dataset~\cite{grauman2022ego4d} has enabled a series of research in visual-language understanding in egocentric videos, 
{\em e.g.}, question answering that focuses on summarizing entire video content~\cite{mangalam2023egoschema}; 
natural language query~(NLQ) that requires temporal localization based on a given query~\cite{ramakrishnan2023naq}; 
grounded question answering that considers answering the query, while localizing the query-related time span simultaneously~\cite{barmann2022did}. However, the above-mentioned settings tend to fall into an over-simplistic scenario, where questions are typically answerable based on visual cues from a single time point, 
or only one time span is annotated among multiple relevant spans. 
For instance, the question \textit{``How many shirts did I pack in my suitcase?"} is deliberately excluded if the \textit{packing} process occurs across multiple, non-contiguous time spans, as described in the annotation process of the NLQ task in Ego4D.

As a consequence, VidQA systems built upon the above-mentioned tasks can hardly be applied in multi-hop scenarios, due to two primary limitations: 
the scarcity of data supporting multi-hop reasoning, 
the deficiency in architecture design to support multi-hop temporal perception. 
{\em First}, there is a notable insufficiency of training data on questions that require reasoning across multiple temporal spans, especially in long-form egocentric videos;
{\em Second}, existing architectures that treat temporal grounding as a language modeling task~\cite{ren2024timechat, huang2024vtimellm, huang2024lita}, 
{\em e.g.}, directly incorporating the timestamp as the target of auto-regressive prediction, which is less effective than task-specific models~\cite{lin2023univtg, mu2024snag}.

To bridge the gap, this paper introduces the problem of Multi-Hop Video Question-Answering \textit{(MH-VidQA)}. As illustrated in Fig.~\ref{fig:teaser}, this task requires the model to simultaneously answer questions that involve visual information from multiple time intervals and localize these time spans as evidence within long, egocentric videos.

To acquire the data necessary for visual instruction tuning, 
we have developed an automated pipeline to construct large-scale question-answer-evidence triplets from the narrations of Ego4D~\cite{grauman2022ego4d}.
Specifically, we build action scene graphs~\cite{ji2020action} by extracting syntax trees from narrations, allowing us to analyze the temporal progression of actions, 
objects, and their relationships, thereby identifying potential questions that require information from multiple time points to answer. We then utilize large language models~(LLMs) to generate eligible triplets across six different question types, that encompass real-world scenarios with an emphasis on the interactions between the individual and the external environment, as well as the long-term temporal relation of events. 
The categories include repeated activities, multiple actions, multiple objects, multiple locations/people, event composition, and event comparison.

Leveraging our automatically constructed data for visual instruction tuning, 
despite the existing VideoLLM~\cite{ren2024timechat} has demonstrated improved multi-hop reasoning abilities, it still struggles to localize the relevant time spans, primarily due to the limitations in predicting timestamps accurately. 
We further propose a novel architecture, termed as \textbf{G}rounding Scattered \textbf{E}vidence with Large \textbf{L}anguage \textbf{M}odel~(\modelname). 
This architecture incorporates grounding tokens into the vocabulary of a multi-modal large language model, that are generated within the responses and then fused with visual features in a temporal grounding module to provide corresponding evidences, 
thereby enhancing the interpretability of the answers. 

To track the development progress on \textit{MH-VidQA} task, 
we have established a new benchmark, termed as \textbf{\smalldatasetname}, 
that involves participants for validating and refining the generated triplets. Comprehensive evaluations show that both proprietary and open-source large multi-modal models largely fall behind human performance, highlighting the substantial challenge presented by \textbf{\smalldatasetname}. Our architecture, trained on the automatically constructed instruction-tuning data, has shown significant improvement in multi-hop reasoning and grounding. We also evaluate our architecture on another public single-hop VidQA benchmark, ActivityNet-RTL~\cite{huang2024lita}, outperforming existing approaches by a large margin.

\begin{table}[!t]
  \centering
  \resizebox{\linewidth}{!}{
    \begin{tabular}{lccccc}
        \toprule
        \multirow{2}{*}{Dataset} & \multirow{2}{*}{Annotation}  & Avg.  & \multirow{2}{*}{Ego?} & Time  & Multi- \\
        & & Duration (s) &  & Labels & Spans \\
        \midrule
        % \specialrule{0em}{1pt}{1pt} 
        \multicolumn{6}{c}{\large\gray{\textit{Conventional\; VidQA\; Benchmarks}}} \\
        \addlinespace[0.2em]
        % \midrule
        MovieQA  & Manual & 211.4 & \rxmark & \gcmark & \rxmark \\
        MSRVTT-QA  & Auto & 15 & \rxmark & \rxmark & \rxmark \\
        MSVD-QA  & Auto & 10 & \rxmark & \rxmark & \rxmark \\
        TVQA  & Manual & 76.2 & \rxmark & \gcmark & \rxmark \\
        % ActivityNet-QA  & Manual & 180 & \rxmark & \rxmark & \rxmark \\
        How2QA  & Manual & 60 & \rxmark & \gcmark & \rxmark \\
        NeXT-QA  & Manual & 44 & \rxmark & \rxmark & \rxmark \\
        iVQA & Manual & 18.6 & \rxmark & \rxmark & \rxmark \\
        EgoSchema & Auto + Manual & 180 & \gcmark & \rxmark & \rxmark \\
        % Perception Test & Manual & 23 & \rxmark & \rxmark & \rxmark \\
        % MovieChat-1K  & Manual & 564 & \rxmark & \rxmark  & \rxmark \\
        % CinePile  & Auto + Manual & 160 & \rxmark & \rxmark & \rxmark \\
        \midrule
        \multicolumn{6}{c}{\large\gray{\textit{Grounded\; VidQA\; Benchmarks}}} \\
        \addlinespace[0.2em]
        % \midrule
        QAEgo4D & Manual & 498 & \gcmark & \gcmark & \rxmark \\
        NExT-GQA & Manual & 39.5 & \rxmark & \gcmark & \rxmark \\
        ActivityNet-RTL & Auto + Manual & 180 & \rxmark & \gcmark & \rxmark \\
        % \midrule
        % \specialrule{0em}{0.3pt}{0.3pt} 
        \rowcolor{gray!15}
        \hrulespace\textbf{\tabledatasetname} & Auto + Manual & 180 & \gcmark & \gcmark & \gcmark \\ %
        \bottomrule
    \end{tabular}
}
  \vspace{-5pt}
  \caption{
  \textbf{Comparison of VidQA benchmarks.} 
  Our proposed benchmark focuses on assessing multi-hop reasoning and grounding abilities within long-form egocentric videos.
  }
  \label{dataset_comparison_table}
  \vspace{-0.3cm}
\end{table}

%% file: sections/2_related_work.tex
\section{Related Work}

\noindent \textbf{Video Question Answering Datasets.} 
Video Question Answering (VidQA) is a video understanding task that involves answering natural language queries using visual-only or multi-modal information from videos. MovieQA~\cite{tapaswi2016movieqa} proposes one of the earliest datasets in this area. However, most of its questions can be answered based on subtitles alone, with few relying on visual cues~\cite{jasani2019we}. 
ActivityNet-QA~\cite{yu2019activitynetqa} and How2QA~\cite{sanabria2018how2} have focused on visual understanding in daily life and instructional videos. 
More recent datasets like
NeXT-QA~\cite{xiao2021nextqanext}, Perception Test~\cite{patraucean2024perception}, STAR~\cite{wu2021star}, and AGQA~\cite{grundemclaughlin2021agqa} focus on designing questions requiring spatio-temporal reasoning and causal relations. Additionally, EgoSchema~\cite{mangalam2023egoschema} proposes to generate questions through LLMs and manual efforts for long-form egocentric videos.

\vspace{3pt} \noindent \textbf{Multi-Hop QA with Grounding.} 
In Natural Language Processing, 
multi-hop question-answering involves reasoning across multiple pieces of information, often requiring the retrieval of evidence from various sources~\cite{yang2018hotpotqa, ho2020constructing, xiong2021answering, trivedi2022musique, zhang2024end}.
In video understanding, conventional VidQA benchmarks do not necessitate models to explicitly localize or reason over temporally scattered evidence. However, recent works like \textsc{\normalsize E\small go\normalsize T\small ime\normalsize QA}~\cite{di2024grounded}, NExT-GQA~\cite{xiao2024can}, and \textsc{\normalsize R\small E\normalsize X\normalsize T\small I\small M\small E}~\cite{chen2024rextime} emphasize the importance of the grounding evidence in VidQA.
These benchmarks, though, assume that evidence is confined to a single time span, overlooking the need for long-term temporal modelling and multi-step reasoning,  which can be an oversimplification in video understanding. %\revised{In this paper, we consider the problem of \textit{MH-VidQA} and explore trustworthy VidQA systems.}

\begin{figure*}[!th]
    \centering
    \includegraphics[width=0.98\linewidth]{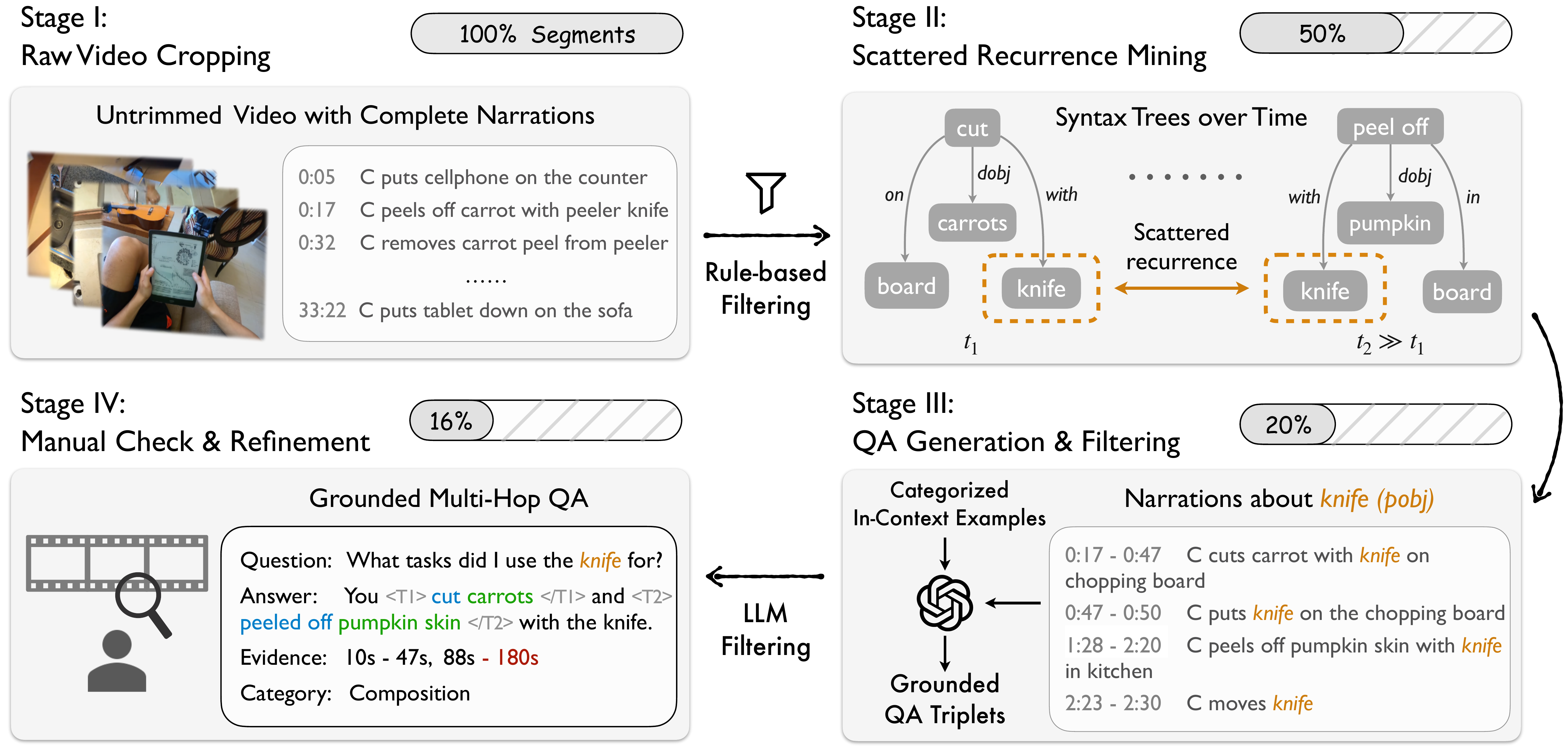}
    \vspace{-0.1cm}
    \caption{\textbf{Illustration of our data curation pipeline.} To collect large-scale multi-hop VidQA data, we have developed an automated pipeline. We begin by using action scene graphs to identify potential multi-hop reasoning questions based on the syntax trees of annotated narrations. Next, we use $\mathtt{GPT\,\text{-}\,4o}$ to generate data samples that include questions, answers, and relevant time spans. Finally, we perform manual validation and refinement to create the new benchmark.}
    \label{fig:curation}
    \vspace{-0.2cm}
\end{figure*}

\vspace{3pt} \noindent \textbf{Multi-modal Large Language Models.}
With the recent advancements in Large Language Models (LLMs)~\cite{achiam2023gpt, vicuna2023, llama3modelcard, jiang2024mixtral}, researchers are endeavouring to develop Multi-modal Large Language Models (MLLMs) by aligning visual and linguistic modalities through visual instruction tuning. 
For image understanding, several studies~\cite{alayrac2022flamingo, li2023blip, zhu2023minigpt, liu2023llava, liu2024llavanext} have shown strong performance across various VQA benchmarks~\cite{lu2023mathvista, liu2023mmbench, yue2024mmmu}. 
In video understanding, while some works~\cite{mvbench2023, minigpt4video-2024, zhang2024llavanextvideo} have made progress on traditional VidQA benchmarks, they are generally designed for short videos. 
Recent efforts to improve temporal awareness~\cite{ren2024timechat, huang2024vtimellm, qian2024momentor} still lag behind task-specific models~\cite{lin2023univtg, mu2024snag} in the temporal grounding ability. 
To address this gap, our proposed dataset supports both instruction tuning and the evaluation of multi-hop reasoning and grounding in long-form egocentric videos, thereby advancing the development of video-language models.

%% file: sections/3_curation.tex
\section{Problem Formulation}
Given a video stream and a question in the format of free-form text,
{\em e.g.}, $\mathcal{V}$ and $\mathcal{Q}$ respectively, 
the objective is to generate the answer and localise the temporal evidence :
\begin{equation}
[\mathcal{\hat{A}}, \mathcal{\hat{T}}] = \Phi (\mathcal{V}, \mathcal{Q}),
\end{equation}
where $\mathcal{\hat{T}} = \{[s_1, e_1], [s_2, e_2], \dots, [s_n, e_n]\}$ refers to a set of non-overlapping start-end time intervals, in which the video content is necessary for deriving the answer $\mathcal{\hat{A}}$.

To develop the vision systems that address our considered \textit{MH-VidQA} task, 
it is essential to collect data in triplet form, {\em i.e.}, $(\mathcal{Q, A, T})$, to train the architecture that can simultaneously answer questions, 
and ground them across multiple time spans.
In the following sections, we will detail an automated pipeline for constructing visual instructions and training our proposed architecture.

\section{\bigdatasetname: Curation Pipeline}
The curation pipeline involves four stages: 
(i) cropping and selecting video clips from untrimmed Ego4D dataset; 
(ii) mining potential questions that demand multi-hop reasoning based on narrations; 
(iii) producing $(\mathcal{Q, A, T})$ triplets across various categories using LLMs; 
(iv) filtering the generated samples with LLMs, then followed by manual review and refinement.

\subsection{Raw Video Cropping \& Selection}
We start with the 9,611 untrimmed egocentric videos~(24-minute duration on average), accompanied with a total of 3.85M timestamped narrations from Ego4D~\cite{grauman2022ego4d}. 
Since these timestamps indicate the occurrence of a new action (\ie, the start time), to estimate the duration of each action, we take the timestamp of the subsequent narration as the end time.
Specifically, we segment the raw videos into non-overlapping 3-minute clips. Each clip and the corresponding narrations are denoted as $\mathcal{V}$ and $\mathcal{N}=\{N_i\}_{i=1}^{\lvert \mathcal{N} \rvert}$.

\subsection{Mining Multi-Hop QA from Narrations}
We propose to mine the multi-hop VidQA triplets from action scene graphs for each long-form video, which provide temporally evolving object descriptions, human-object relationships, 
and the progression of actions over time.~\cite{ji2020action, yang2023pvsg, rodin2024action}.

To build action scene graphs, we use the syntax tree of each narration to identify the specific nodes, involving actions, objects, locations and people. We then search for structures where a single node recurs over time, but connects with different neighbouring nodes across various scenes, since these structures are likely to contain the multi-hop reasoning queries. The detailed procedure is outlined below.

\vspace{3pt} \noindent 
\textbf{Narration Syntax Tree $\rightarrow$ Action Scene Graph.} 
We use \texttt{spaCy}~\cite{honnibal2020spacy} to parse each narration and extract the basic nodes, 
including \action{actions} \textit{(verb)}, \object{direct objects} \textit{(dobj)}, and \location{prepositional objects} \textit{(pobj)} along with their modifiers. 
For instance, ``C puts the cooking pot on the counter top'' 
can be parsed into \{\action{put}, \object{cooking pot}, \location{counter top}\} as three nodes with distinct syntactic attributes.

% \begin{figure}[!ht]
%     \centering
%     \includegraphics[width=0.9\linewidth]{figs/syntax_tree.pdf}
%     \caption{Simplified syntax tree of a certain narration.}
%     \label{fig:syntax_tree}
% \end{figure}

\vspace{3pt} \noindent 
\textbf{Searching Scattered Recurrence in Graph.}
We then focus on a node $u$ that recurs sporadically throughout the entire set of narrations $\mathcal{N}$.
The minimum and maximum recurrence times for the selected node $u$ are denoted as $t_{min}$ and $t_{max}$, respectively, which determines the number of time intervals involved in the question.
We extract the narrations related to the specific node $u$ from $\mathcal{N}$, denoted as $\mathcal{N}_{u}$. 
These narrations, though focused on node $u$, describe different action scenes, making them potential candidates for multi-hop triplet generation in Stage III.

\subsection{QA Generation \& Filtering with LLM}
Based on the nodes' syntactic attributes, we use different in-context learning examples to guide the LLM-based QA generation processes.
The resulting multi-hop questions are divided into six categories, involving repeated activities, multiple actions, multiple objects, multiple locations/people, event composition, and event comparison. The detailed prompts and question examples are presented in the Supplementary Material. Formally, for the selected narrations $\mathcal{N}_u$, the triplet is generated with prompt~($\mathcal{P}$), 
denoted as:
\begin{equation}
    \mathcal{\{(Q, A, T)}\} = \operatorname{LLM} (\mathcal{P};\mathcal{N}_u)
\end{equation}
After the automated generation, we use an LLM to filter out unreasonable QA pairs, resulting in 4,412 clips with 14,397 triplets. We utilize $\mathtt{GPT\,\text{-}\,4o}$ for both generation and filtration processes due to its superior capabilities.

\subsection{Manual Check \& Refinement}

To construct a benchmark, we select 380 clips with 1,208 triplets and hire 12 graduate students majoring in computer vision, to validate the clarity of the data and further refine the temporal annotations. As a result, we obtain 360 clips with 1,080 triplets, which form the final benchmark, termed as \textbf{\smalldatasetname}. The annotation details and benchmark statistics are provided in the Supplementary Material.

% 330 + 51 -> 314 + 46
% the efficiency of our automatic pipeline and the quality of instruction-tuning split

%% file: sections/4_architecture.tex
\section{GeLM: A Baseline Method for \textit{MH-VidQA}}

\begin{figure*}[!th]
    \centering
    \includegraphics[width=0.96\linewidth]{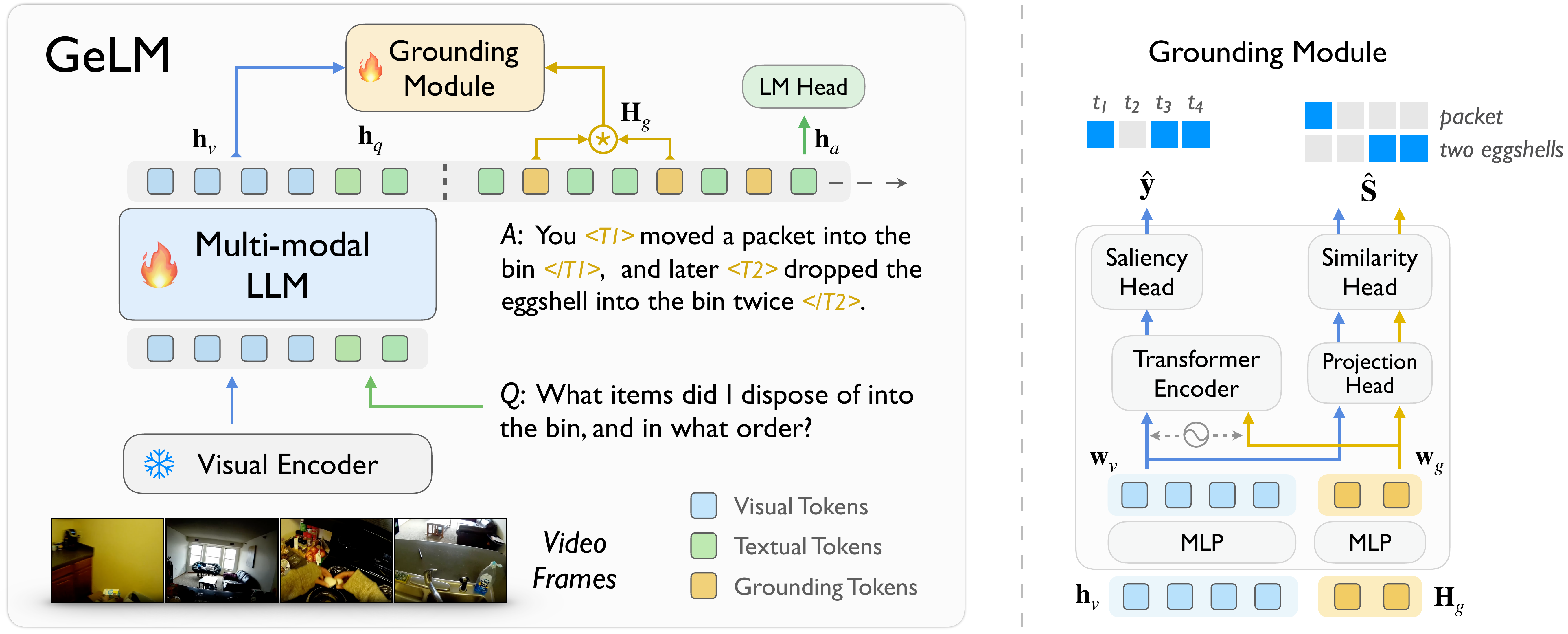}
    \vspace{-0.05cm}
    \caption{\textbf{Overview of the proposed architecture.} \modelname~can generate grounding token pairs, \ie, \gtokens, in the response of a multi-modal large language model, which denote the start and end times of the enclosed statement. These grounding tokens are then processed with visual hidden states to the ground multiple time spans that provide evidence supporting the answer.}
    \label{fig:arch}
    \vspace{-0.1cm}
\end{figure*}

Existing models for video question answering typically provide answers without supporting temporal evidence, or are restricted to identifying a single time interval. 
Here, we propose a novel architecture, termed as \textbf{GeLM}: \textbf{G}rounding Scattered \textbf{E}vidence with Large \textbf{L}anguage \textbf{M}odel for Multi-Hop Video Question-Answering. As depicted in Fig.~\ref{fig:arch}, 
our model primarily comprises a multi-modal large language model and a grounding module, with special grounding tokens~(\gtokens\hspace{1pt}) indicating the time span of the enclosed key information in the response. 

\subsection{Visual-language Encoding Module}
Given the video clip with $L$ frames and the associated question, we first adopt a frozen visual encoder to extract the visual features. The given question is tokenized and transformed into textual embeddings, while the visual features are projected into visual embeddings with the same dimension through a linear projector:
\begin{equation}
    \mathbf{x}_v = \phi_{\text{proj}}(\Phi_{\text{v-enc}}(\mathcal{V})), \quad \mathbf{x}_q = \phi_{\text{emb}}(\mathcal{Q}) 
\end{equation}
where $\mathbf{x}_v \in \mathbb{R}^{L \times D}$, 
$\mathbf{x}_q \in \mathbb{R}^{Q \times D}$ denote the computed embeddings for visual and question respectively.

\subsection{Multi-modal Large Language Model}
The visual and textual embeddings are then fed into a multi-modal large language model:
\begin{equation}
   \{\mathbf{h}_v, \mathbf{h}_q, \mathbf{h}_a\} = \operatorname{MLLM}([\mathbf{x}_v : \mathbf{x}_q])
\end{equation}
where $\mathbf{h}_v, \mathbf{h}_q, \mathbf{h}_{a}$ represent the hidden states of the input frames, question, and the output response respectively. The answer texts $\mathcal{\hat{A}}$ are then decoded using a linear head on $\mathbf{h}_{a}$.

\vspace{4pt} \noindent \textbf{Grounding Tokens.}
Inspired by approaches that enable MLLMs to segment visual entities~\cite{lai2024lisa, zhang2024groundhog, yan2024visa}, 
we expand the vocabulary by adding grounding token pairs, \ie, \tstart{} \tend{}, which indicate the start-end time span. 
As illustrated in Fig.~\ref{fig:arch}, when the MLLM needs to ground the temporal evidence for its response, the relevant part of the response is enclosed by \tstart{} and \tend{}. 
% \weidi{the following sentences are difficult to understand.}
We concatenate the last-layer hidden states of each pair, 
\ie, \tstart{1} and \tend{1}, $\ldots$, \tstart{K} and \tend{K} along the channel dimension to form a single grounding query vector, resulting in $K$ grounding queries $\mathbf{H}_g \in \mathbb{R}^{K\times 2D}$. Note that, the value of $K$ can vary for different responses.
These queries are then processed through the grounding module, which interacts with the visual hidden states $\mathbf{h}_v \in \mathbb{R}^{L\times D}$.

\subsection{Evidence Grounding Module}
To ground the time spans that support the answer, 
we design an evidence grounding module that processes a variable number of grounding queries and predicts the corresponding temporal proposals in the video: $\mathcal{\hat{T}} = \{[s_i, e_i]\}_{i=1}^{\lvert \mathcal{\hat{T}} \rvert}$.

We begin by projecting the hidden states of frames $\mathbf{h}_v$ and grounding queries $\mathbf{H}_g$ into the same dimension $C$:
\begin{equation}
    \mathbf{w}_v = \phi_v (\mathbf{h}_v) \in \mathbb{R}^{L\times C}, \quad \mathbf{w}_g = \phi_g (\mathbf{H}_g) \in \mathbb{R}^{K\times C}
\end{equation}
Following this, two separate branches are used to predict the temporal evidence for the answer: the saliency branch and the similarity branch, as depicted in Fig.~\ref{fig:arch}.
The saliency branch utilizes a self-attention mechanism across all visual features and grounding tokens to identify all temporal evidence for the question holistically. 
The similarity branch calculates the visual-textual similarity between each grounding query and all visual features, to determine the time spans for each part of the response in a fragmented manner.

\vspace{4pt} \noindent \textbf{Saliency branch.} 
As illustrated in Fig.~\ref{fig:arch}, the saliency branch utilizes three Transformer Encoder layers as the temporal aggregator, to fuse information between grounding queries and visual hidden states:
\begin{equation}
   \{\mathbf{o}_v, \mathbf{o}_g\}  = \Phi_{\text{temp-agg}}([\mathbf{w}_v : \mathbf{w}_g])
\end{equation}
The predicted saliency score $\mathbf{\hat{y}}\in\mathbb{R}^{L}$ is derived through the saliency head $\phi_{\text{saliency}}(\cdot)$ and the sigmoid function $\sigma(\cdot)$: 
\begin{equation}
    \mathbf{\hat{y}} = \sigma(\phi_{\text{saliency}}(\mathbf{o}_v))
\end{equation}
where a higher score indicates a higher probability that each frame serves as visual evidence for the question-answering. The saliency head $\phi_{\text{saliency}}(\cdot)$ consists of two Conv1D layers with ReLU activation.

\vspace{4pt} \noindent \textbf{Similarity branch.} 
Apart from predicting the saliency, we also aim to determine the time spans of key information enclosed by each grounding token pair separately, represented as a similarity matrix $\mathbf{\hat{S}}\in\mathbb{R}^{K \times L}$ between $K$ grounding queries and $L$ frames. Specifically, we project the hidden states with a linear layer, 
\begin{equation}
    \mathbf{z}^v = \psi_{\text{v-proj}}(\mathbf{w}_v), \quad \mathbf{z}^g = \psi_{\text{g-proj}}(\mathbf{w}_g)
\end{equation}
% \weidi{insert the notation into the text description}
and compute the cosine similarity matrix:
\begin{align}
    \mathbf{\hat{S}}_{ij} = \frac{{\mathbf{z}^g_i} \cdot {\mathbf{z}^v_j}}{\left\lVert \mathbf{z}^g \right\rVert \cdot \left\lVert \mathbf{z}^v \right\rVert} \in [-1, 1]
\end{align}
where a higher value of value of $\mathbf{\hat{S}}_{ij}$ indicates that the $j$-th frame is more relevant to the $i$-th grounding query.

\begin{table*}[t]
% \small
\centering
\resizebox{\linewidth}{!}{
\begin{tabular}{l>{\centering\arraybackslash}cm{2cm}cccccc}
\toprule
\specialrule{0em}{2pt}{2pt}
\multirow{2}{*}{\largehrulespace\textbf{Methods}} & \multirow{2}{*}{\largehrulespace\textbf{Mode}} & \multirow{2}{*}{\largehrulespace\textbf{Input}}  & \multicolumn{4}{c}{\textbf{Temporal Grounding}} & \multicolumn{2}{c}{\textbf{Question Answering}}  \\
\specialrule{0em}{0.5pt}{0.5pt}
\cmidrule(r){4-7} \cmidrule(r){8-9}
\specialrule{0em}{0.5pt}{0.5pt}
~ & ~ & ~ & mIoP   & mIoG & \textbf{IoU@0.3}  & \textbf{mIoU}   & Sent. Sim. & \textbf{Score} $(10\uparrow)$ \\
\specialrule{0em}{1pt}{1pt} 
\hline
\specialrule{0em}{2pt}{2pt} 
Human & - & - &  71.8  &  81.0  &  87.0  & 61.8  & 74.3  & 7.5    \\
\addlinespace[0.15em]
\arrayrulecolor{gray}
\hdashline
\specialrule{0em}{0.5pt}{0.5pt} 
\largehrulespace GPT-4o~\cite{gpt4o} & \textit{z.s.} & 60 frames  & 18.9  & 24.4  & \cellcolor{CarnationPink!15}12.0 & \cellcolor{CarnationPink!15}12.2  & \cellcolor{CarnationPink!15}73.7 & \cellcolor{LightGreen!15}\textbf{5.4}   \\
\addlinespace[0.25em]
\multicolumn{9}{l}{\gray{\textit{{End-to-End MLLMs}}}}\\ \hrulespace
InternVL2-8B~\cite{internvl-1.5-2024} & \textit{z.s.} & 30 frames &  11.8  &  24.0  & 6.3  &  6.6   & 71.9 & 4.5    \\
\addlinespace[0.05em]\hrulespace
LLaVA-NeXT-Video-7B~\cite{zhang2024llavanextvideo} & \textit{z.s.} & 32 frames & -  &  -  &  - &  -  & 62.1 & 4.2   \\
\addlinespace[0.05em]\hrulespace
TimeChat-7B~\cite{ren2024timechat} & \textit{z.s.} & 96 frames  &  10.2  &  5.6  &  3.0 & 3.6 & 58.9 & 3.3  \\
\addlinespace[0.05em]\hrulespace
VTimeLLM-7B~\cite{huang2024vtimellm} & \textit{z.s.} & 100 frames &  12.4   & \cellcolor{CarnationPink!15}28.2 & 8.8 & 9.2 & 70.5 & 4.3  \\
\addlinespace[0.25em]
\multicolumn{9}{l}{\gray{\textit{{Pipeline: Caption Module $\rightarrow$ LLM (QA + Grounding)}}}} \\ \hrulespace
LLaVa-NeXT-7B $\rightarrow$ Llama-3.1-8B & \textit{z.s.} & 180 frames &  \cellcolor{CarnationPink!15}21.4  & 22.3 & 10.1  &  9.7  & 
63.6 & 3.5  \\
\addlinespace[0.2em]
\arrayrulecolor{gray}\hdashline
\specialrule{0em}{0.5pt}{0.5pt} 
\largehrulespace
\textbf{\modelname-7B} (Ours) & \textit{f.t.} & 180 frames & \cellcolor{LightGreen!15}\textbf{23.7}  & \cellcolor{LightGreen!15}\textbf{41.0} & \cellcolor{LightGreen!15}\textbf{18.2} & \cellcolor{LightGreen!15}\textbf{16.7} & \cellcolor{LightGreen!15}\textbf{75.0} & \cellcolor{CarnationPink!15}4.8 \\
\bottomrule
 \end{tabular}
}
\vspace{-5pt}
\caption{\textbf{Performance of various multi-modal models on \smalldatasetname.} The \colorbox{LightGreen!20}{\textbf{best}} and \colorbox{CarnationPink!15}{second-best} performances of the metrics are highlighted. \textit{`z.s.'} and \textit{`f.t.'} refer to zero-shot and fine-tuning, respectively. Existing approaches of various types fall short of human performance on this challenging task. To bridge this gap and set a new baseline, we have trained our proposed architecture using automatically constructed visual instruction tuning data.
} 
\label{tab:zero-shot}
\vspace{-0.1cm}
\end{table*}

\vspace{4pt} \noindent \textbf{Proposal generation strategy.}
During inference, 
to generate temporal proposals~($\mathcal{\hat{T}}$), 
we apply the following post-processing.
Utilizing the saliency score vector $\mathbf{\hat{y}}\in\mathbb{R}^{L}$, 
we set a threshold at $70\%$ of the maximum saliency score. Timestamps with scores above this threshold are merging into time spans.

Leveraging the similarity matrix $\mathbf{\hat{S}}\in\mathbb{R}^{K \times L}$, we apply an average pooling kernel with a size of 3 and a stride of 1 to smooth the values. Then we perform a softmax function along each row to get positive scores. 
Since each row vector $\mathbf{\hat{S}}_{k,:}\in\mathbb{R}^{L}$ in the matrix represents the predicted temporal relevance for the $k$-th grounding query, we apply the same thresholding method to each row and take the union of the results to obtain a set of proposals.

\vspace{4pt} \noindent \textbf{Training objective.} 
For question answering, the cross entropy loss $\mathcal{L}_{\text{CE}}(\mathcal{\hat{A}}, \mathcal{A})$ is utilized for next token prediction.
For evidence grounding, given the ground truth binary saliency labels $\mathbf{y} \in \{0, 1\}^{L}$, we use binary cross entropy as loss function: $\mathcal{L}_{\text{BCE}} = \frac{1}{L}\sum_{i=1}^{L}-\mathbf{y}_i\log \mathbf{\hat{y}}_i$.
With the ground truth binary similarity matrix $\mathbf{S}\in \{0, 1\}^{K \times L}$, 
we adopt the Multiple Instance Learning NCE (MIL-NCE) loss~\cite{miech19howto100m} for contrastive learning:
\begin{equation}
    \mathcal{L}_{\text{NCE}}= -\frac{1}{K}\sum^K_{i=1}\log \frac{\sum_{j=1}^{L} \mathbf{S}_{ij}\exp(\mathbf{\hat{S}}_{ij}/\tau)}{\sum_{j=1}^{L}\exp(\mathbf{\hat{S}}_{ij}/\tau)}
\end{equation}
where $\tau$ denotes temperature. $i, j$ correspond to $i$-th grounding query and $j$-th frame, respectively. The final loss is a weighted sum of the above losses: $\mathcal{L} = \mathcal{L}_{\text{CE}} + \lambda_{\text{BCE}}\mathcal{L}_{\text{BCE}} + \lambda_{\text{NCE}}\mathcal{L}_{\text{NCE}}$.

%% file: sections/5_experiments.tex
\section{Experiments}
In this section, we first describe the metrics for our benchmark, \textbf{\smalldatasetname}, and then evaluate the performance of existing approaches. Next, we employ instruction tuning with the automatically constructed dataset, to establish a strong baseline for the multi-hop VidQA task. Lastly, we show that our method also achieves state-of-the-art performance on the existing public single-hop VidQA task.

\subsection{Evaluation Metrics}
We evaluate the performance of question answering and evidence grounding separately on \textbf{\smalldatasetname}. 

\vspace{3pt} \noindent \textbf{Question answering.} 
To evaluate open-ended answers, we use GPT-4o as the primary evaluator for scoring, as it more closely aligns with human judgment and is widely adopted for assessment purposes~\cite{chiang-lee-2023-large, zheng2024judging}. 
We also report the average Sentence Similarity (Sent. Sim.) between the ground truth answers and the predicted answers~\cite{reimers2019sentence}.
For time-related questions, we exclude them from metrics of answering, as they can be accurately evaluated with localization metrics.

\vspace{3pt} \noindent \textbf{Evidence localization.} 
Given the $m$ predicted non-overlap time spans: 
$\mathcal{\hat{T}} = \{\hat{T}_1, \hat{T}_2, \ldots, \hat{T}_m\}$ and the ground truth consisting of $n$ spans: $\mathcal{T}=\{T_1, T_2, \dots, T_n\}$ for each video, IoU (Intersection over Union) is computed as follows:
\begin{equation}
    \text{IoU}(\mathcal{T}, \hat{\mathcal{T}}) = \frac{\sum_{i=1}^{m} \sum_{j=1}^{n} |\hat{T}_i \cap T_j|}{\left| \bigcup_{i=1}^{m} \hat{T}_i \cup \bigcup_{j=1}^{n} T_j \right|}
    \label{eq:iou}
\end{equation}
This can be seen as an extension of the IoU between two intervals, measuring the Jaccard Distance between two sets of time spans.
Subsequently, the mean IoU (mIoU) is determined by averaging the IoU values across the entire test set. 
Additionally, we calculate the proportion of videos with an IoU exceeding 0.3, designated as IoU@0.3.
Similar to precision and recall, we compute mIoP and mIoG by averaging Intersection over Prediction (IoP) and Intersection over Ground Truth (IoG), replacing the denominator in Eq.~\eqref{eq:iou} with \scalebox{0.9}{$\left| \bigcup_{i=1}^{m} \hat{T}_i\right|$} and \scalebox{0.9}{$\left|\bigcup_{j=1}^{n} T_j \right|$}, respectively.
% \cite{xiao2024can}

% Due to the varying levels of descriptive granularity present in open-ended responses to the same question, we do not employ Hungarian Matching between \(m\) predicted intervals and \(n\) ground truth intervals \revised{(\eg, there maybe two short time intervals for the response ``wash one green bowl and wash one yellow bowl", and only one long interval for the ground truth ``wash two bowls", but both of them are plausible to the question ``What did I wash?''). }

\subsection{Evaluation on \datasetname}
In this section, we evaluate several latest multi-modal models on \textbf{\smalldatasetname}, exploring their abilities of multi-hop reasoning and temporal grounding.

\vspace{3pt} \noindent \textbf{Human and advanced proprietary model.} 
Initially, we invite participants~(different from annotators in the curation pipeline) to assess human performance on this task. 
We randomly sample 10\% of the test split and request participants to answer the questions and localise relevant time spans.
Additionally, we evaluate the advanced proprietary model, 
GPT-4o, by leveraging its visual capabilities with uniformly sampled frames from the video clip, to perform answering and grounding.

\vspace{3pt} \noindent \textbf{End-to-end models.} 
We conduct investigations across various popular MLLMs, 
including Image LLM (InternVL2-8B), Short Video LLM (LLaVA-NeXT-Video-7B), and Long Video LLMs (TimeChat-7B, VTimeLLM-7B).

\vspace{3pt} \noindent \textbf{Multi-stage pipeline.} 
To explore the effectiveness of dense captioning for \textit{MH-VidQA}, we adopt a multi-stage pipeline, 
consisting of an image caption module, followed by an LLM. 
The captions of sampled frames with timestamps will be utilized by the LLM for answering and grounding.

\vspace{4pt} \noindent \textbf{Overall Results.}
From experiments presented in Tab.~\ref{tab:zero-shot}, we can draw the following observations: 
\emph{1) Both the proprietary model and open-source multi-modal LLMs significantly lag behind human performance}, 
underscoring the current limitations in multi-hop reasoning and grounding capabilities within multi-modal systems.
\emph{2) Reasoning and grounding abilities are disentangled in existing visual systems.} For instance, LLaVA-NeXT-Video is unable to handle requests involving temporal grounding, but can still answer part of questions that do not involve temporal grounding.
\emph{3) Instruction-tuning with single-hop data does not guarantee superiority in multi-hop grounding.} For example, despite TimeChat and VTimeLLM have been fine-tuned with temporally aware instructions and multi-turn conversations, the ability to ground multiple intervals for a single query remains limited.
\emph{4) Dense captions do indeed help temporal grounding, but errors may cascade.} Although captioning at per second provides explicit temporal information for grounding, errors in the captioning process are difficult to correct through the subsequent stages.
\textbf{We recommend that readers refer to the evaluation details and additional qualitative results in the Supplementary Material.}

\subsection{A Baseline Method for \datasetname}
In the following section, we propose a new baseline for this challenging task and conduct ablation experiments to evaluate the effectiveness of the automatically constructed training data, and our architectural design for future research.

\subsubsection{Implementation Details}
\paragraph{Training data.} 
We utilize the triplets generated in our automated pipeline to train the multi-modal LLM and the grounding module. 
These triplets have been filtered by the LLM, 
but not manually refined in Stage IV, 
consisting of 3,156 clips with a total of 10,414 samples.

\vspace{3pt} \noindent \textbf{Architecture.} 
The visual features of \smalldatasetname\hspace{1pt} are extracted with the InternVideo-MM-L-14~\cite{wang2022internvideo} from 8 frames per second. The large language model employed is Vicuna-7B v1.3~\cite{vicuna2023}. The dimensions of the hidden states for the LLM and the grounding module are 4096 and 1024, respectively.

\vspace{3pt} \noindent \textbf{Training setup.}  
The experiments are conducted using 4 NVIDIA H800 (80GB) GPUs, with a batch size of 32 per device. 
The model is trained for 10 epochs with a learning rate of $2 \times 10^{-5}$, employing a warmup cosine decay strategy.

\subsubsection{Ablation Studies}

\begin{table}[!t]
% \small
\centering
\resizebox{\linewidth}{!}{
\begin{tabular}{lcccc}
\toprule
\specialrule{0em}{2pt}{2pt}
\multirow{2}{*}{\hrulespace\textbf{Training Loss}} & \multirow{2}{*}{\hrulespace\textbf{Strategy}} & \multicolumn{2}{c}{\textbf{Temporal Grounding}} & \multicolumn{1}{c}{\textbf{QA}}  \\ 
\specialrule{0em}{0.3pt}{0.3pt}
\cmidrule(r){3-4} \cmidrule(r){5-5}
\specialrule{0em}{0.1pt}{0.2pt}
 &  & \textbf{IoU@0.3}  & \textbf{mIoU}  & \textbf{Score} $\uparrow$ \\ 
\midrule
 \hrulespace $\mathcal{L}_{\text{CE}}$  & -  &  -  &  -  & 4.7 \\ [0.2em]
\arrayrulecolor{gray}\hdashline
\specialrule{0em}{2pt}{2pt}
 $+ \mathcal{L}_{\text{BCE}}$  & Saliency   & 13.8   &  14.2  & 4.7 \\ [0.1em]
 \addlinespace[2.5pt]
 $+ \mathcal{L}_{\text{NCE}}$  & Similarity  & 14.1   &  13.4  & 4.6 \\ [0.1em]
\largehrulespace \multirow{2}{*}{$+ \mathcal{L}_{\text{NCE}} + \mathcal{L}_{\text{BCE}}$}  & Saliency  &  \textbf{19.2}  &  14.7  & 4.7 \\ [0.1em]
\hrulespace   & \cellcolor{gray!10}Similarity  &  \cellcolor{gray!10}18.2  &  \cellcolor{gray!10}\textbf{16.7}  & \cellcolor{gray!10}\textbf{4.8} \\[0.01em]
\bottomrule
\end{tabular}
}
\vspace{-5pt}
\caption{\textbf{Ablation of the training objective and inference strategy.} The \colorbox{gray!10}{gray} shading indicates the default setting.}
\label{abalation:arch}
\vspace{-0.4cm}
\end{table}

\paragraph{Effect of training objective and inference strategy.}
We explore the role of each training loss and the effect of the two branches on generating temporal proposals. 
As shown in Tab.~\ref{abalation:arch}, 
although the saliency branch is unable to distinguish the time interval of each grounding token pair, the binary cross entropy loss tends to benefit the temporal grounding, 
improving the performance of the similarity branch, with IoU@0.3 increasing from 14.1 to 18.2, and mIoU from 13.4 to 16.7.
Correspondingly, the similarity branch also enhances the inference results of the saliency branch, demonstrating the complementarity of both branches.

\vspace{4pt} \noindent \textbf{Effect of the visual instruction-tuning data.} 
To validate the effectiveness of our data curation pipeline for mining large-scale multi-hop VidQA data, we utilize the automatically collected instructions to fine-tune a pre-trained Video LLM, \eg,~TimeChat, and evaluate on \smalldatasetname. 
As Tab.~\ref{abalation:datasize} shows, visual instruction tuning enhances the multi-hop reasoning and grounding abilities of TimeChat, demonstrating the effectiveness of constructed data.
Additionally, we trained our \modelname\hspace{1pt} model by varying percentages of data, while maintaining the same number of iterations to explore the effect of the data scale. 
The continuous performance improvement from the increased data volume demonstrates the potential of our automated pipeline to collect large-scale data and enhance model capabilities.

\begin{table}[!t]
% \small
\centering
\resizebox{\linewidth}{!}{
\begin{tabular}{ccccc}
\toprule
\specialrule{0em}{2pt}{2pt}
\multirow{2}{*}{\hrulespace \textbf{Method}} & \multirow{2}{*}{\hrulespace \textbf{Training Data}} & \multicolumn{2}{c}{\textbf{Temporal Grounding}} & \multicolumn{1}{c}{\textbf{QA}}  \\ 
\specialrule{0em}{0.3pt}{0.3pt}
\cmidrule(r){3-4} \cmidrule(r){5-5}
\specialrule{0em}{0.1pt}{0.2pt}
 &  & \textbf{IoU@0.3}  & \textbf{mIoU}  & \textbf{Score} $\uparrow$ \\ 
\midrule
\addlinespace[5pt]
\multirow{2}{*}[-.1em]{\hrulespace TimeChat}  & \ding{55}  &  3.0  &  3.6  & 3.3 \\
\addlinespace[0.1em]
\hrulespace  & 100\% &  8.6  &  8.1  &  4.4 \\
\addlinespace[0.2em]
\arrayrulecolor{gray}\hdashline
\addlinespace[4pt]
\multirow{3}{*}[-.5em]{Ours}  & 25\%  &  13.0  &  11.3  & 4.6 \\
\addlinespace[0.1em]
\hrulespace  & 50\%  &  16.7  & 16.1   & 4.7 \\
\addlinespace[0.1em]
  % \rowcolor{gray!10}
\largehrulespace  & \cellcolor{gray!10}100\%  & \cellcolor{gray!10}\textbf{18.2}  & \cellcolor{gray!10}\textbf{16.7}  & \cellcolor{gray!10}\textbf{4.8} \\
\bottomrule
\end{tabular}
}
\vspace{-3pt}
\caption{\textbf{Effect of the instruction-tuning data.} 
We explore on both the pre-trained model and our proposed architecture.}
\vspace{-0.4cm}
\label{abalation:datasize}
\end{table}

\subsection{On Existing Single-Hop VidQA Benchmark}

\vspace{3pt} \noindent \textbf{Dataset and metrics.}
In addition to our multi-hop benchmark, we validate the effectiveness of our method on the public single-hop VidQA benchmark~\cite{huang2024lita}, which contains 229 question-answer pairs across 160 videos. For this benchmark, the temporal grounding metrics are mIoU and Precision@0.5 (P@0.5), with the latter measuring the percentage of predictions with an IoU over 0.5. Additionally, the GPT-4 Relative Score (R. Score) is computed for evaluating the predicted explanations.

\vspace{3pt} \noindent \textbf{Comparison.}
In the existing state-of-the-art method, for example, LITA~\cite{huang2024lita} adds special time tokens into the vocabulary to process temporal grounding as a next-token prediction task on this benchmark. As shown in Tab.~\ref{tab:rtl}, our architecture significantly exceeds LITA with both temporal grounding branches after fine-tuning. 
\vspace{-0.15cm}

%which integrates a grounding module in the LLM with specific losses designed to explicitly enhance and supervise the temporal grounding process. 

\begin{table}[!ht]
% \small
\centering
\resizebox{\linewidth}{!}{
\begin{tabular}{lclll}
\toprule
\specialrule{0em}{2pt}{2pt}
\multirow{2}{*}{\hrulespace\textbf{Model}} & \multirow{2}{*}{\hrulespace\textbf{Strategy}} & \multicolumn{2}{c}{\textbf{Temporal Grounding}} & \multicolumn{1}{c}{\textbf{QA}}  \\ 
\specialrule{0em}{0.3pt}{0.3pt}
\cmidrule(r){3-4} \cmidrule(r){5-5}
\specialrule{0em}{0.1pt}{0.2pt}
 &  & \multicolumn{1}{l}{\textbf{mIoU}}  & \multicolumn{1}{l}{\textbf{P@0.5}}  &\textbf{R. Score} $\uparrow$ \\ 
\midrule
% \hrulespace Video-LLaMA-v2-13B & - & - & - & 32.1 \\
% \addlinespace[0.1em]
% \hrulespace Video-ChatGPT-13B & - & - & - & 38.8 \\
% \addlinespace[0.1em]
\hrulespace LITA-7B & Time Token & 24.1 & 21.2 & 44.0 \\ 
\addlinespace[0.1em]
\hrulespace LITA-13B & Time Token & 28.6 & 25.9 & 46.3 \\ 
\addlinespace[0.15em]
\arrayrulecolor{gray}\hdashline
\addlinespace[0.05em]
\largehrulespace  Ours-7B  & Saliency &  31.8  & 28.2   & \textbf{45.3} \\ 
\addlinespace[0.05em] 
\rowcolor{gray!10}
\largehrulespace  Ours-7B  & Similarity  &  35.4 \garrow{11.3}  &  \textbf{31.0} \garrow{9.8}  & 45.1 \garrow{1.1} % 51.2
\\
\addlinespace[0.01em]
\bottomrule
\end{tabular}
}
\vspace{-4pt}
\caption{
\textbf{Comparison with the state-of-the-art method on ActivityNet-RTL}, a public single-hop VidQA benchmark.
}
\label{tab:rtl}
\end{table}

%% file: sections/6_conclusion.tex
\section{Conclusion}

To conclude, we have initiated the \textit{MH-VidQA} task for long-form egocentric video understanding. To acquire the associated dataset, 
we have devised an automated pipeline to mine large-scale multi-hop QA triplets, a subset of which are subsequently validated and refined manually, resulting in a new benchmark.
Existing multi-modal systems demonstrate improvement in multi-hop reasoning abilities after training on the automatically collected data, but they still struggle to ground temporal evidence for their responses effectively due to weak temporal perception.
To bridge this gap, we have proposed a novel model capable of answering multi-hop questions and concurrently grounding scattered visual clues, which establishes a baseline for this challenging task after visual instruction tuning. Our method also achieves state-of-the-art performance on the public single-hop VidQA benchmark, further underscoring its effectiveness.

%% file: sections/7_appendix.tex
% \clearpage
% \mysection{Appendix}

\appendix
\onecolumn
{\centering
\LARGE
\textbf{Grounded Multi-Hop VideoQA in Long-Form Egocentric Videos}\\
\Large
\vspace{0.6em}Supplementary Material \\
\vspace{2.0em}
}

\nopagebreak

\section{Benchmark Details}

\vspace{5pt} \noindent This section provides additional statistical details about the proposed benchmark, \textbf{\smalldatasetname}, outlines the question types categorized by human annotators, details the automated pipeline rules and prompts for generating and filtering question-answer-evidence triplets, and introduces the user interfaces for human annotation and participant evaluation.

\subsection{Additional Statistics}

\begin{figure*}[htbp]
    \centering
    \begin{subfigure}[b]{0.48\textwidth}
        \centering
        \includegraphics[width=\textwidth]{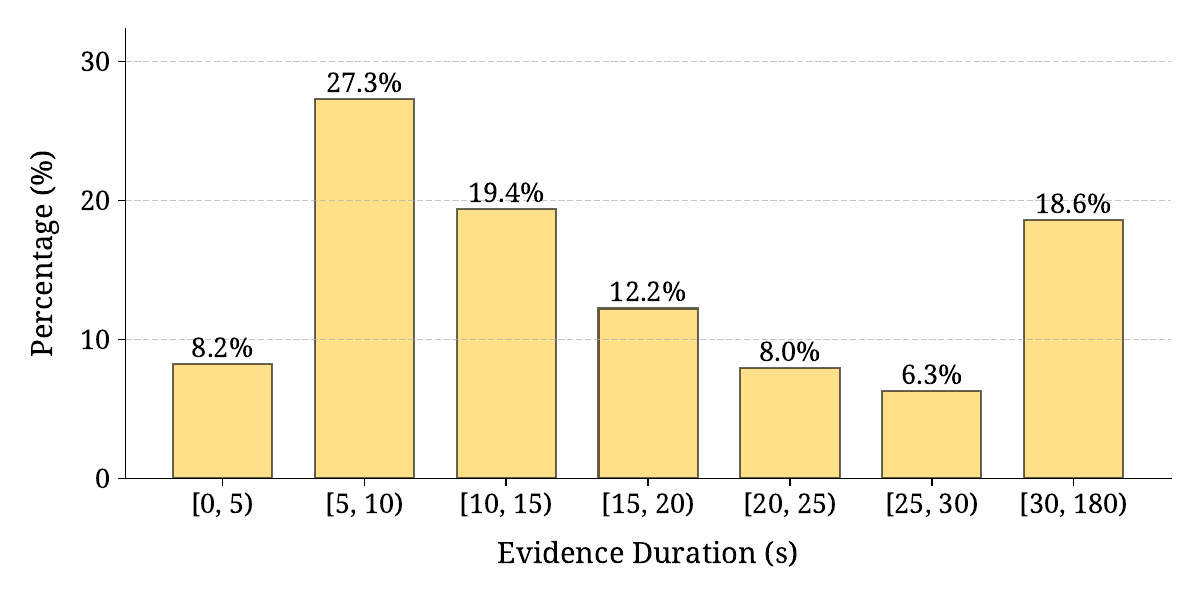}
        \caption{Histogram of temporal evidence duration.}
        \label{fig:statistic-duration}
    \end{subfigure}
    \hfill
    \begin{subfigure}[b]{0.48\textwidth}
        \centering
        \includegraphics[width=\textwidth]{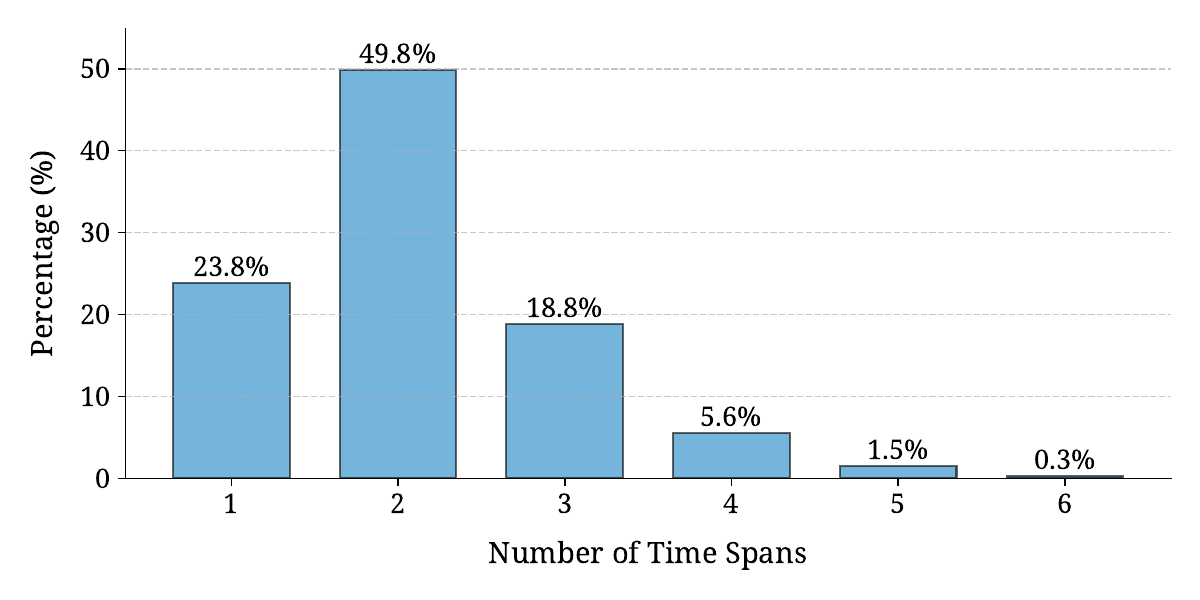}
        \caption{Histogram of time span counts.}
        \label{fig:statistic-hops}
    \end{subfigure}
    
    \vskip 20pt
    
    \begin{subfigure}[b]{0.48\textwidth}
        \centering
        \includegraphics[width=\textwidth]{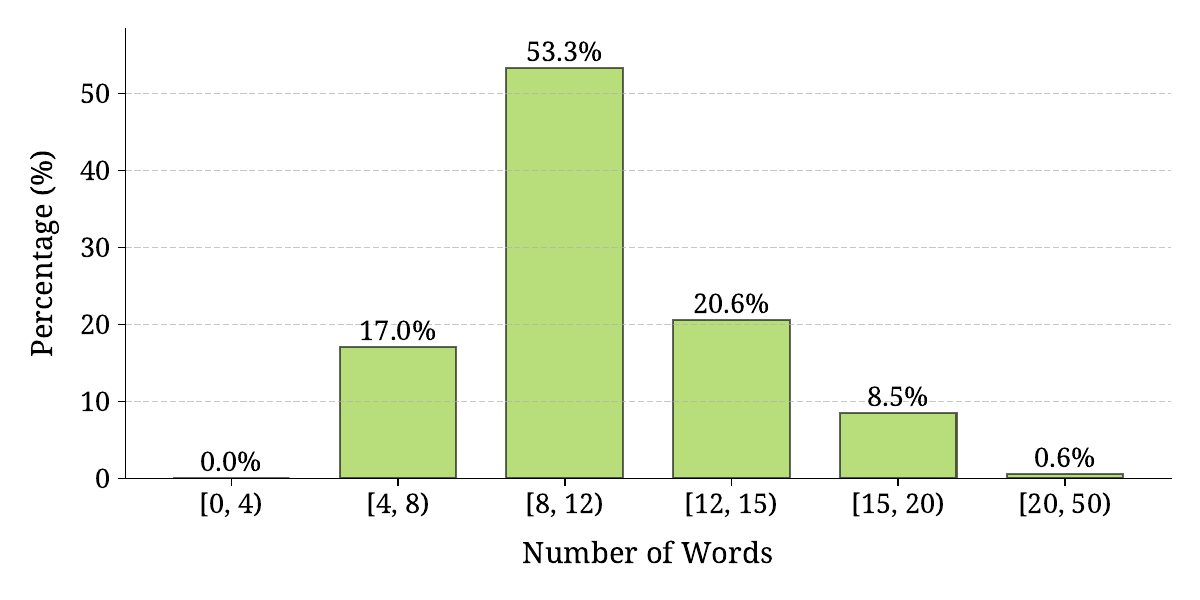}
        \caption{Histogram of question word counts.}
        \label{fig:statistic-q-words}
    \end{subfigure}
    \hfill
    \begin{subfigure}[b]{0.48\textwidth}
        \centering
        \includegraphics[width=\textwidth]{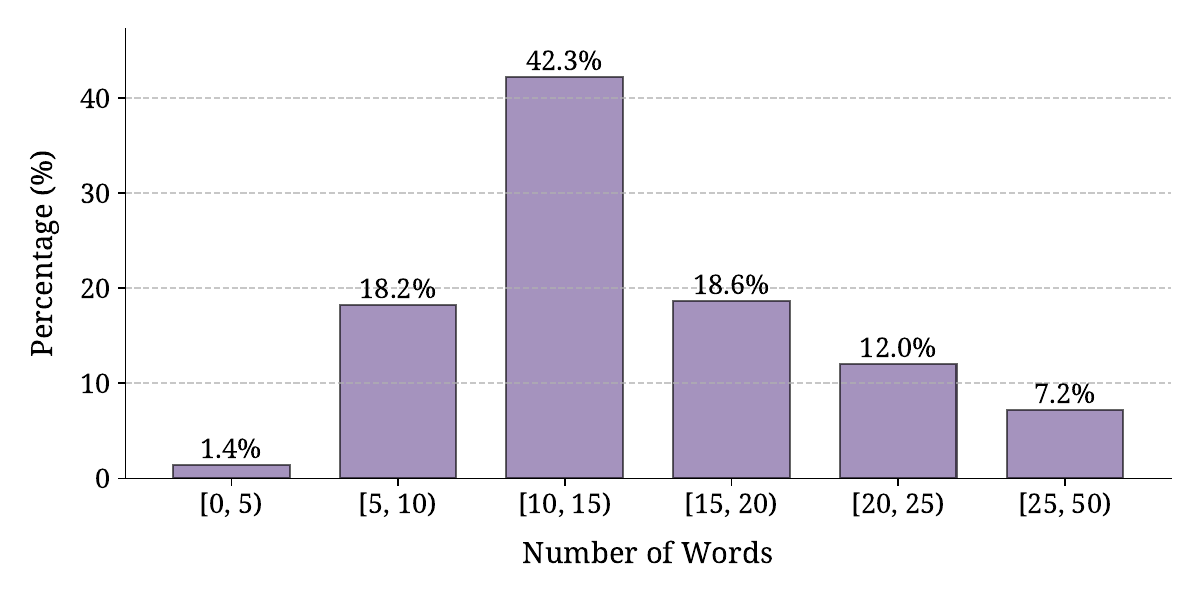}
        \caption{Histogram of answer word counts.}
        \label{fig:statistic-a-words}
    \end{subfigure}
    
    \caption{\textbf{Visualization of \smalldatasetname~statistics.}}
    \label{fig:statistics}
\end{figure*}

As illustrated in Figure~\ref{fig:statistics}, we present a statistical analysis of \textbf{\smalldatasetname}, encompassing the duration of temporal evidence, the number of time spans involved in the questions, and the distribution of word counts in both questions and answers. 
Our calculations reveal that the average duration of temporal evidence is 19.5 seconds, with half of the instances lasting less than 15 seconds. 
The average number of time spans, or ``hops'', is 2.1, with a maximum of 6 hops observed. 
Furthermore, the average word counts for questions and answers are 10.3 and 14.4, respectively, reflecting a notable level of complexity.

\vspace{5pt}

\subsection{Multi-Hop Question Categories}

As shown in Table~\ref{tab:category}, 
the questions in \textbf{\smalldatasetname}~are manually categorized into six groups: repeated activities, multiple actions, multiple objects, multiple locations/people, event composition, and event comparison. 
Notably, all these visual questions require gathering information from multiple time intervals and reasoning across them to provide accurate answers. 
The temporal distribution of these time spans can be scattered and distant. 
The question format follows the NLQ task style in Ego4D but covers a broader range of scenarios.

\begin{table*}[!th]
    \centering
    \resizebox{0.95\textwidth}{!}{
    \begin{tabular}{
    >{\raggedright\arraybackslash}m{1.5cm}
    >{\raggedright\arraybackslash}m{3.9cm}
    >{\raggedright\arraybackslash}m{6cm}
    >{\raggedright\arraybackslash}m{6cm}@{}
    }
        \toprule
        \textbf{ID} & \textbf{Category} & \textbf{Question} & \textbf{Answer} \\
        \midrule
         A~\graysmall{$(39.6\%)$} &  Repeated Activities & How many times did I open the fridge? & You \tstart{1} opened the fridge \tend{1} three times. \\\addlinespace[0.1em] \thinhline
        
        \hrulespace B~\graysmall{$(22.4\%)$}  & \hrulespace Multiple \action{Actions} & \hrulespace What item did I take out of the fridge and later put back during the video?  & \hrulespace You \tstart{1} \action{took out} \object{a conical flask} \tend{1} and then \tstart{2} \action{put it back} \tend{2}. \\\addlinespace[0.1em] \thinhline
        
        \hrulespace C~\graysmall{$(17.9\%)$}   & \hrulespace Multiple \object{Objects} & \hrulespace What items did I rinse besides my hand? & \hrulespace You \action{rinsed} \tstart{1} \object{the carrot} \tend{1}, \tstart{2} \object{the tomato} \tend{2}, and \tstart{3} \object{the potato peeler} \tend{3}. \\\addlinespace[0.1em] \thinhline
        
        \multirow{2}{*}[-0.8em]{D~\graysmall{$(6.7\%)$}}  & \multirow{2}{*}[-0.8em]{Multiple Locations / People} &  Where did I put the phone? & \hrulespace You put the phone \tstart{1} on the kitchen counter \tend{1} and later \tstart{2} on the dining table \tend{2}. \\\addlinespace[0.1em] %\cline{2-3}
         & &  How many people did I talk to? & \hrulespace Two. You talked to \tstart{1} lady X \tend{1} and \tstart{2} lady Y \tend{2}. \\\addlinespace[0.1em] \thinhline
         
        \hrulespace E~\graysmall{$(10.3\%)$}  & \hrulespace Event Composition & \hrulespace What did I do after opening the box? & \hrulespace You \tstart{1} \action{put} \object{the toys} in\tend{1} , and the second time \tstart{2} \action{took out} \object{candles} \tend{2}. \\\addlinespace[0.1em] \thinhline
        
        \hrulespace F~\graysmall{$(3.1\%)$}  & \hrulespace Event Comparison & \hrulespace Did I wash my hands first or rinse the washing pad? & \hrulespace No. You \tstart{1} rinsed the washing pad \tend{1} between \tstart{2} washing your hands twice \tend{2}. \\
        \bottomrule
    \end{tabular}
    }
    \vspace{-3pt}
    \caption{Examples of each category in \smalldatasetname. Each pair of special tokens \tstart{}\textcolor{gray}{$\ldots$}\tend{}~represents the time intervals of the enclosed referent. `Event Composition' refers to questions that involve both multiple \action{\textbf{actions}} and multiple \object{\textbf{objects}}.}
    \label{tab:category}
\end{table*}

\subsection{Details of the Automated Pipeline}

\paragraph{Rule-based Filtering.}
As described in the main paper, we filter video clips based on the following criteria: (i) those with the number of narrations greater than $60$ or smaller than $30$, (ii) those where the time span between the first narration and last narration is less than $150s$.
This filtering ensures that the narrations of clips are detailed and have balanced temporal granularity.
We select nodes $u$ with recurrence times smaller than $t_{\text{max}}=5$ and greater than $t_{\text{min}}=2$.
Additionally, the time span of the selected narrations $\mathcal{N}_u$ should exceed $10s$ to ensure the distinctness of multiple time intervals.

\paragraph{The prompt for LLM-based Generation and Filtration.}
In Table~\ref{tab:gpt4o-generation} and Table~\ref{tab:gpt4o-filtration}, we present the prompts used for generating and filtering multi-hop VidQA triplets, utilizing \texttt{gpt-4o-2024-05-13}, respectively.
In practice, we employ three different generation prompts for the node $u$, corresponding to the attributes \action{\textit{verb}}, \object{\textit{dobj}}~(direct objects), and \location{\textit{pobj}}~(prepositional objects) .
We empirically design seven in-context learning examples, 
to encompass as many specific forms of the selected narration $\mathcal{N}_u$ about the node $u$ as possible.

\subsection{Details of the Annotation Procedure}

\paragraph{Annotation Interfaces.}

\begin{figure*}[!ht]
    \centering
    \begin{subfigure}[b]{0.92\linewidth}
        \centering
        \includegraphics[width=\linewidth]{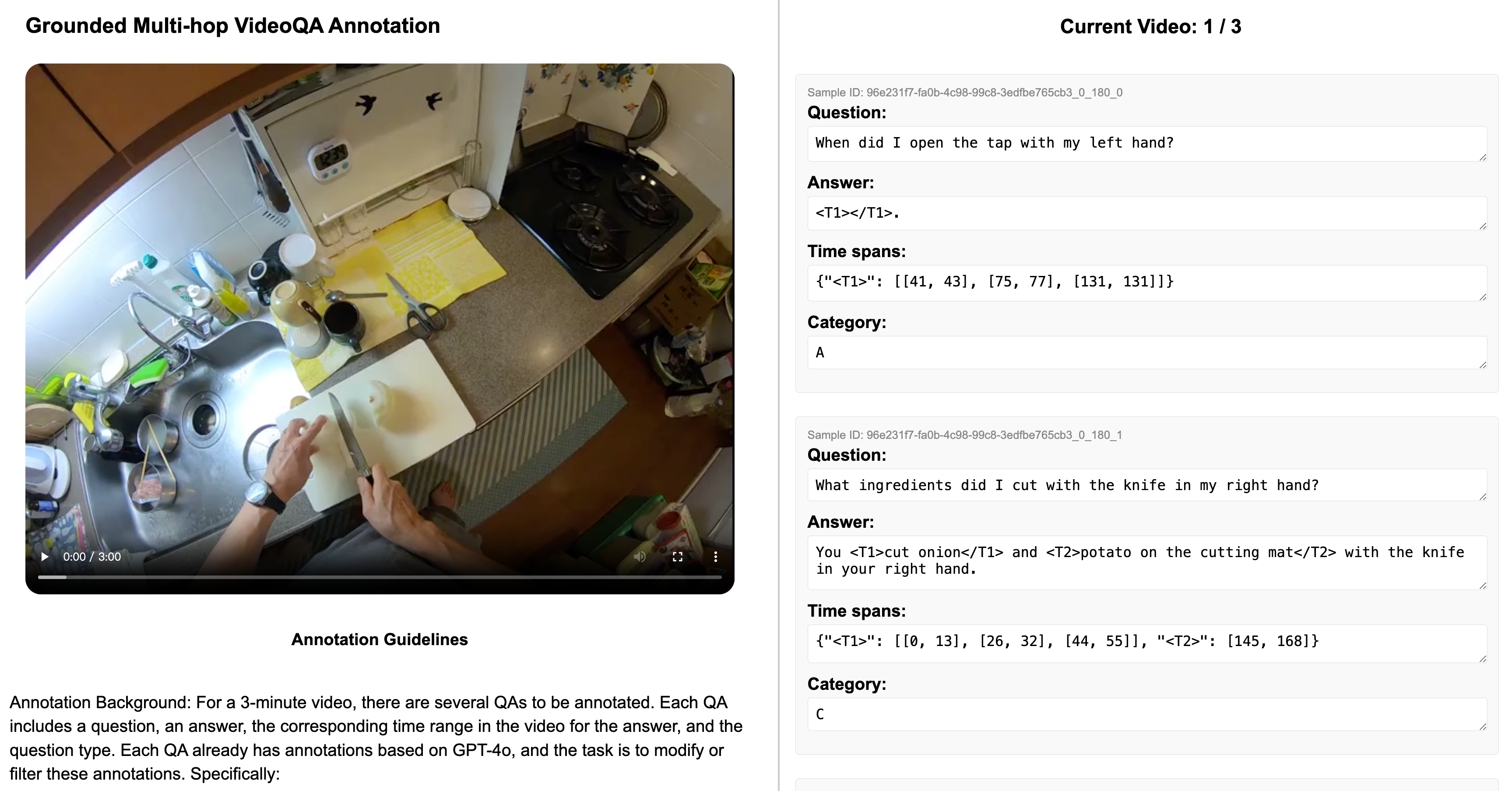}
        \caption{Human annotation interface.}
        \label{fig:interface-annotation}
    \end{subfigure}
    
    \vspace{2em} 

    \begin{subfigure}[b]{0.92\linewidth}
        \centering
        \includegraphics[width=\linewidth]{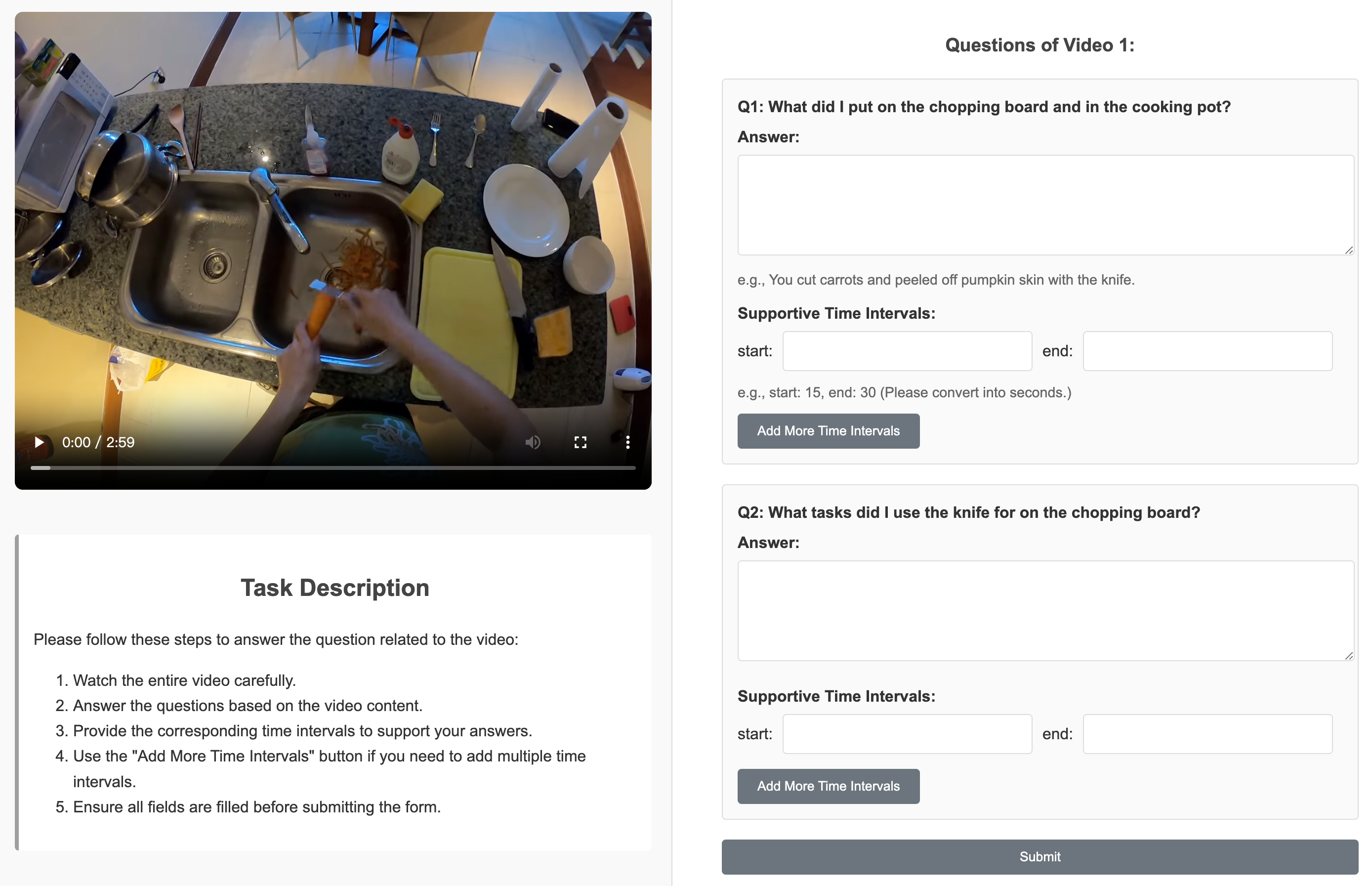}
        \caption{Human evaluation interface.}
        \label{fig:interface-evaluation}
    \end{subfigure}
    \vspace{2pt}
    \caption{User interfaces for the manual annotation and the evaluation of human performance on \smalldatasetname.}
    \label{fig:interfaces}
\end{figure*}

In Figure~\ref{fig:interface-annotation}, we present the annotation interface developed for human annotators to validate the triplets generated and filtered by $\mathtt{GPT\,\text{-}\,4o}$, as well as to refine the time intervals for reasonable question-answer pairs.
Figure~\ref{fig:interface-evaluation} shows the interface designed for participants different from previous annotators, to watch the video clip with the accompanying questions. 
These participants are required to answer the question and localise the time spans that support their answers. The results are then used to assess the human performance of answering and grounding on the Multi-Hop VidQA task.

\paragraph{Annotation Guidelines.} The detailed guidelines, as illustrated in the interface of Figure~\ref{fig:interface-annotation}, are outlined as follows:

\vspace{5pt}
\begin{mdframed}
\textit{Annotation Background:} For a 3-minute video, there are several QAs to be annotated. Each QA includes a question, an answer, the corresponding time range in the video for the answer, and the question type. Each QA already has annotations based on GPT-4o, and the task is to modify or filter these annotations. Specifically:

\begin{enumerate}
    \item For each QA, first check if the question and answer are reasonable based on the video content. 
    \item Each key action or information mentioned in the answer should be enclosed with a pair of delimiters \texttt{<T></T>}, representing the time interval it occurs in the video. Note that each action may appear in a single time period \([s, e]\) or repeatedly in multiple time periods \(\left[ [s_1, e_1], \ldots [s_n, e_n] \right]\) (units in seconds, \(0 \leq s \leq e \leq 179\)).
    \item Annotate the type of QA based on the examples in the table (A, B, … , F). If there is a significant discrepancy between the question and answer, mark it directly as U, representing `Unusable'.
    \item When the question directly asks for the timing of an event (When did I …?), the answer should always be \texttt{<T></T>}. For this case, only ensure that the temporal annotation is correct.
\end{enumerate}

Thank you for participating!
\end{mdframed}
\vspace{5pt}

\begin{table*}[!ht]
\centering
\begin{minipage}{0.95\linewidth} % columnwidth
    \vspace{0mm} \centering
    \begin{tcolorbox}%[colback=lightgray!15, colframe=black!50, coltext=black]
        \centering
        \footnotesize
        \begin{tabular}{@{}p{\linewidth} c}
            \VarSty{ {\bf System Prompt} } & \\
            \addlinespace[0.1em]
            You are a helpful assistant designed to output JSON. & \\
            \hrulefill & \\
            \addlinespace
            \VarSty{ {\bf User:} } & \\
            I want you to act as a teacher in a class called `Video Understanding'. I will provide video action descriptions with timestamps, and your task is to generate one question-and-answer (QA) pair for your students.  & \\
            - Create questions that require integrating information from more than one provided action description to answer. & \\
            - Ensure the question is specific and concrete, not general. Ensure the question does not contain any timestamp. & \\
            - Ensure the answer is undisputedly right to your question. Provide concise sentences for the answer. & \\
            - Enclose the mentioned action or objects in your answer between two paired special tokens (e.g., \textless T1\textgreater~and \textless /T1\textgreater, \textless T2\textgreater~and \textless /T2\textgreater, ...), and next output the time span of each enclosed information according to descriptions with timestamps. & \\
            - When the input action descriptions are similar (like only the hand used is different), directly ask about the time spans of the action (When did I ...?). The answer should include only special tokens when directly asking about time spans (e.g., Example 4). & \\
            - Merge the same, similar or continuous action in your answer and their time span. (e.g., Example 3). & \\
            \addlinespace[8pt]
            Here are examples of the input descriptions with timestamps, and the output QA pair: & \\
            \addlinespace[5pt]
            Example 1: & \\
            \addlinespace
            User: & \\
            start, end, description & \\
            92, 94, C talks with lady B. & \\
            96, 97, C talks with lady B. & \\
            167, 173, C talks with lady X. & \\
            176, 179, C talks with lady X. & \\
            \addlinespace
             Assistant: & \\
            \{`Question': `Who did I talk with?', & \\
              \hspace{5pt}`Answer': `You \textless T1\textgreater~talked with two ladies including \textless T1\textgreater~lady B \textless /T1\textgreater~and \textless T2\textgreater~lady X \textless /T2\textgreater.', & \\
              \hspace{5pt}`Time span':\{`\textless T1\textgreater': [92, 97], `\textless T2\textgreater': [167, 179]\}\} & \\
            \addlinespace[10pt]
            Example 2: & \\
            \addlinespace
            User: & \\
            start, end, description & \\
            22, 23, C places the syringe on the table. & \\
            27, 28, C places the container in the machine. & \\
            149, 150, C places the syringe on the counter. & \\
            163, 174, C places the container on the counter. & \\
            \addlinespace
            Assistant: & \\
            \{`Question': `Where did I place the syringe on during the video?', & \\
              \hspace{5pt}`Answer': `You \textless T1\textgreater~placed the syringe on the table~\textless /T1\textgreater~and \textless T2\textgreater~on the counter~\textless /T2\textgreater.', & \\
              \hspace{5pt}`Time span':\{`\textless T1\textgreater': [22, 23], `\textless T2\textgreater': [149, 150]\}\} & \\
            \addlinespace[8pt]
            % Example 3: & \\
            % \addlinespace
            % User: & \\
            % start, end, description & \\
            % 33, 35, C removes dirt from brush with his left hand. & \\
            % 61, 63, C removes pan support with his right hand. & \\
            % 68, 69, C removes pan support with his right hand. & \\
            % \addlinespace
            % Assistant: & \\
            % \{`Question': `Did I remove pan support first or remove dirt from brush?', & \\
            %   \hspace{5pt}`Answer': `You \textless T1\textgreater~removed pan support\textless /T1\textgreater~after \textless T2\textgreater~removing dirt from brush\textless /T2\textgreater.', & \\
            %   \hspace{5pt}`Time span':\{`\textless T1\textgreater': [61, 69], `\textless T2\textgreater': [33, 35]\}\} & \\
            % \addlinespace[8pt]
            Example 3: & \\
            $\ldots$ & \\ 
            \addlinespace[10pt]
            Now, it's your turn and let's think step by step. Based on the following descriptions with timestamps to generate one required QA pair:  & \\
            \addlinespace[8pt]
            User: & \\
            start, end, description & \\
            \textcolor{teal}{\textless start\textgreater, \textless end\textgreater, \textless narration\textgreater} & \\
            \textcolor{teal}{$\ldots$} & \\ 
            \addlinespace[5pt]
             \VarSty{ {\bf Assistant:} } & \\
             \addlinespace[0.05em]
        \end{tabular}
    \end{tcolorbox}
    \vspace{-5pt}
    \caption{Prompt for $\mathtt{GPT\,\text{-}\,4o}$ to generate multi-hop VidQA triplets. This prompt is designed for the \action{\textbf{action}} node (\eg, \action{\textbf{talk}} of example 1, and  \action{\textbf{place}} of example 2). The remaining five examples are omitted due to space constraints.}
    \label{tab:gpt4o-generation}
\end{minipage}
\end{table*}

\begin{table*}[!ht]
\centering
\begin{minipage}{0.95\linewidth} % columnwidth
    \vspace{0mm} \centering
    \begin{tcolorbox}%[colback=lightgray!15, colframe=black!50, coltext=black]
        \centering
        \footnotesize
        \begin{tabular}{@{}p{\linewidth} c}
            \VarSty{ {\bf System Prompt} } & \\
            \addlinespace[0.1em]
            You are a helpful assistant designed to output JSON. & \\
            \hrulefill & \\
            \addlinespace
            \VarSty{ {\bf User:} } & \\\addlinespace[0.05em]
            I will provide a QA pair that tests students' understanding of an egocentric video, focusing on activities featured in the video. The question require student to integrate information from more than one time span to answer. Your task is to evaluate whether the QA is reasonable. Specifically, analyze if any of the following issues are present:  & \\
            1. The question is too vague, ambiguous or general. & \\
            2. Given the answer, the question is not natural or well-defined. & \\
            3. The question involves ambiguous actions like `what did I touch,' which can have various degrees of interpretation (e.g., holding vs. briefly touching). The similar verbs include `check', `inspect', `adjust', etc. & \\
            \addlinespace
            For each input QA, you first need to output whether it is reasonable, and then provide the rationale for your judgement. In your output, 0 indicates the QA is reasonable, and 1 indicates the QA is unreasonable. & \\
            \addlinespace[5pt]
            Examples of reasonable QAs: & \\
            \addlinespace
            Example 1: & \\
            Input: \{`Q': `Where did I place the syringe during the video?', `A': 'You placed the syringe on the table and on the counter.'\} & \\
            Output: \{`Judgement': 0, `Rationale': `There are no obvious problems. The answer can be determined only by watching the video.'\} & \\
            \addlinespace
            Example 2: & \\
            Input: \{`Q': `Did I remove the pan support first or remove dirt from the brush?', `A': `You removed the pan support after removing dirt from the brush.'\} & \\
            Output: \{`Judgement': 0, `Rationale': `The question and answer are clear and specific, allowing for a well-defined response based on the video content.'\} & \\
            \addlinespace
            Example 3: & \\
            Input: \{`Q': `What liquid did I pour on the sponge?', `A': `You poured water and detergent on the sponge.'\} & \\
            Output: \{`Judgement': 0, `Rationale': `The question is well-defined and specific. The answer can be determined only by watching the video and is possibly right to the question.'\} & \\
            \addlinespace
            Example 4: & \\
            $\ldots$ & \\ 
            \addlinespace[12pt]
            Examples of unreasonable QAs: & \\
            \addlinespace
            Example 1: & \\
            Input: \{`Q': `What did I remove before the pan support?', `A': `You removed dirt from the brush before removing the pan support.'\} & \\
            Output: \{`Judgement': 1, `Rationale': `Removing dirt and removing the pan support are two distinct actions despite both involving the verb `remove.' Grouping them in one question makes it difficult for students to formulate the correct answer.'\} & \\
            \addlinespace
            Example 2: & \\
            Input: \{`Q': `What objects did I open?', `A': `You opened a file, a door, and a bucket.'\} & \\
            Output: \{`Judgement': 1, `Rationale': `The question is too broad and vague regarding the verb `open.' It is challenging for students to categorize opening a door and opening a file as the same type of action.'\} & \\
            \addlinespace
            % Example 3: & \\
            % Input: \{`Q': `Where did I search?', `A': `You searched in the file and in the bucket.'\} & \\
            % Output: \{`Judgement': 1, `Rationale': `The question seems to ask about a location, but the answer focuses on specific objects. A student might answer `in the room' and still be correct, indicating that the question is ambiguous.'\} & \\
            % \addlinespace
            Example 3: & \\
            $\ldots$ & \\ 
            \addlinespace[12pt]
            Now, it's your turn and let's think step by step. Based on the following descriptions with timestamps to generate one required QA pair:  & \\
            \addlinespace[5pt]
            Input: \textcolor{teal}{\textless QA Sample\textgreater} & \\
            \addlinespace[8pt]
             \VarSty{ {\bf Assistant:} } & \\\addlinespace[0.05em]
        \end{tabular}
    \end{tcolorbox}
    \vspace{-5pt}
    \caption{Prompt for $\mathtt{GPT\,\text{-}\,4o}$ to preliminarily filter the unreasonable multi-hop VidQA samples, before further manual validation and refinement. The remaining examples are omitted due to space constraints.}
    \label{tab:gpt4o-filtration}
\end{minipage}
\end{table*}

\clearpage

\section{Evaluation Details}

\vspace{5pt} \noindent In this section, we begin by introducing the baseline models which are evaluated in \textbf{\smalldatasetname}, including multi-modal models and a multi-stage pipeline. Subsequently, we provide the prompts and detailed settings employed in the inference process with these systems. 
Furthermore, we report their QA performance on additional open-ended answering metrics, some of which are not utilized in \textbf{\smalldatasetname}, as they may be deemed unsuitable for evaluating long-form answers.

\subsection{Baselines and Inference Settings}

\paragraph{GPT-4o~\cite{gpt4o}.} 
We uniformly sample one frame every three seconds from the video clip and leverage the visual perception abilities of \texttt{gpt-4o-2024-05-13} to simultaneously answer the given question and ground the supporting evidence. 
Additionally, we employ regular expressions to extract time intervals from the response. The `detail' parameter for the API function is set to `low'. The prompt used is presented in Table~\ref{tab:zs-gpt-4o}.

\paragraph{InternVL2~\cite{internvl-1.5-2024}.} 
InternVL2 is an open-source Multi-modal Large Language Model (MLLM), which supports diverse input modalities and multitask outputs. We utilize the multi-turn conversational ability of InternVL2 to assess its performance in both answering and grounding. Leveraging its capability of multi-image input, we uniformly sampled 30 frames from the video clip.
The specific checkpoint we used is \texttt{OpenGVLab/InternVL2-8B} from Hugging Face. The prompt used is presented in Table~\ref{tab:zs-internvl2}.

\paragraph{LLaVA-NeXT-Video~\cite{zhang2024llavanextvideo}.} 
LLaVA-NeXT-Video is a model that excels in video understanding tasks, leveraging techniques like AnyRes for representing high-resolution images and length generalization for handling long videos.
We utilize the checkpoint \texttt{llava-hf/LLaVA-NeXT-Video-7B-hf} from Hugging Face for evaluation.
Despite testing multiple prompts, LLaVA-NeXT-Video encounters difficulties in generating time-related responses. 
As a result, we report only the question-answering metrics for this model. 
The detailed prompt is presented in Table~\ref{tab:zs-llava-next-video}.

\paragraph{TimeChat~\cite{ren2024timechat}.} 
TimeChat is a multimodal large language model designed for long video understanding, featuring a timestamp-aware frame encoder and a sliding video Q-Former, supported by the TimeIT dataset with 125K instances for improved instruction-following. 
Given that TimeChat has designed specific instruction templates for temporal grounding, we leverage its multi-turn conversational capabilities to first answer the question and then follow its predefined prompt styles to ground the answer. The number of input frames is 96 as default.
The detailed prompt is presented in Table~\ref{tab:zs-llava-next-video} and the checkpoint is downloaded from the official GitHub repository.

\paragraph{VTimeLLM~\cite{huang2024vtimellm}.} 
VTimeLLM is a Video LLM designed for precise video moment understanding and temporal reasoning, employing a three-stage training strategy, namely, feature alignment, temporal-boundary awareness, and enhancement of temporal understanding.
We adhere to the multi-turn conversational templates of VTimeLLM, starting with question answering, then followed by grounding the response. 
By default, 100 frames are sampled as visual input. 
The detailed prompt is provided in Table~\ref{tab:zs-vtimellm}.

\paragraph{Multi-stage Pipeline.} 
To construct a multi-stage pipeline for multi-hop VidQA, 
we first employ an image captioning module, \texttt{llava-hf/llava-v1.6-mistral-7b-hf} from Hugging Face, to generate captions for frames uniformly sampled from the video per second, with prompts shown in Table~\ref{tab:zs-pipeline-caption}. 
This process yields 180 captions per video. 
For the reasoning module, we utilize \texttt{meta-llama/Meta-Llama-3.1-8B-Instruct}. The prompts used are similar to those employed for GPT-4o, as shown in Table~\ref{tab:zs-pipeline-reason}, with the distinction being the input, which consists of either video frames or frame captions.

\subsection{Additional Metrics for Evaluating Open-Ended Responses}

\begin{table*}[!ht]
\centering
\resizebox{0.9\linewidth}{!}{
\begin{tabular}{l>{\centering\arraybackslash}m{2.5cm}ccccccc}
\toprule
\specialrule{0em}{2pt}{2pt}
\multirow{2}{*}{\largehrulespace\textbf{Methods}} & \multirow{2}{*}{\largehrulespace\textbf{Input}} & \multicolumn{7}{c}{\textbf{Question Answering}}  \\ 
\specialrule{0em}{0.5pt}{0.5pt}
\cmidrule(r){3-9}
\specialrule{0em}{0.5pt}{0.5pt}
~ & ~ & BLEU-4 & METEOR & ROUGE-L & CIDEr & Sent. Sim & \textbf{Score} $\uparrow$ \\
\specialrule{0em}{1pt}{1pt} 
\hline
\specialrule{0em}{2pt}{2pt} 
% Human & - &   &   &   &   & 74.3  & 7.5  \\[0.15em]
GPT-4o & 60 frames & 17.9 & 21.8 & 42.8 & 172.6 & 73.7 & 5.4 \\[0.1em]
\arrayrulecolor{gray}\hdashline
\specialrule{0em}{1pt}{1pt} 
\multicolumn{8}{l}{\gray{\textit{{End-to-End MLLMs}}}}\\[0.15em]
InternVL2-8B & 30 frames & 21.0 & 24.1 & 43.6 & 173.6 & 71.9 & 4.5 \\[0.15em]
LLaVA-NeXT-Video-7B & 32 frames & 8.6 & 21.3 & 32.0 & 78.1 & 62.1 & 4.2 \\[0.15em]
TimeChat-7B & 96 frames & 10.3 & 16.3 & 32.0 & 79.7 & 58.9 & 3.3 \\[0.15em]
VTimeLLM-7B & 100 frames & 20.9 & 24.0 & 47.3 & 176.8 & 70.5 & 4.3  \\[0.15em]
\arrayrulecolor{gray}\hdashline
\specialrule{0em}{1pt}{1pt} 
\multicolumn{8}{l}{\gray{\textit{{Pipeline: Caption Module $\rightarrow$ LLM (QA + Grounding)}}}} \\[0.2em]
LLaVa-NeXT-7B $\rightarrow$ Llama-3.1-8B & 180 frames & 6.4 & 16.0 & 36.8 & 123.3 & 63.6 & 3.5 \\
\bottomrule
 \end{tabular}}
    \vspace{-4pt}
    \caption{Zero-shot performance comparison on additional open-ended question answering metrics.}
    \label{tab:ans_metrics}
\end{table*}

As introduced in the main text, we utilize $\mathtt{GPT\,\text{-}\,4o}$ as the primary evaluator to score the open-ended responses based on the given questions and the corresponding ground truth answers.
The specific prompt used for evaluation is presented in Table~\ref{tab:gpt4o-evaluate}.

Prior studies~\cite{barmann2022did} adopt BLEU-4~\cite{papineni2002bleu}, METEOR~\cite{banerjee2005meteor} and ROUGE~\cite{lin2004rouge} as metrics for evaluating short answers within 5 words, which are not suitable for long answers typical of our benchmark. 
In Table~\ref{tab:ans_metrics}, we report the zero-shot performance of multi-modal models on these evaluation metrics.
It is important to note that, `zero-shot' means the models have not been explicitly fine-tuned on the training set of \textbf{\smalldatasetname}, though Ego4D videos or other egocentric videos might be involved in training some of the evaluated models.
We observe that only Sentence Similarity~(Sent. Sim.) aligns with the judgements of GPT-4o, compared with BLEU-4, METEOR, and ROUGE-L. 
CIDEr tends to favour responses that paraphrase the question before answering.
Specifically, we utilize \texttt{all-MiniLM-L6-v2} from the Sentence Transformers library to extract sentence embeddings, following~\cite{di2024grounded}.

\vspace{15pt}

\begin{table*}[!ht]
\centering
\begin{minipage}{0.95\linewidth} % columnwidth
    \vspace{0mm} \centering
    \begin{tcolorbox}%[colback=lightgray!15, colframe=black!50, coltext=black]
        \centering
        \footnotesize
        \begin{tabular}{@{}p{\linewidth} c}
            \VarSty{ {\bf User:} } & \\
            \addlinespace[2pt]
            Here is a 3-minute egocentric video recording my activities. I will provide you with frames sampled every 3 seconds from the video. Your task is to answer a specific question based on these frames accurately. Ensure your answer is concise. & \\
            After answering, list the time intervals related to the question as evidence. Each interval should have a start and end timestamp in seconds (e.g., $[[9, 15], [120, 135]]$). Note that the duration of the video is 180s and the frames are sampled at 0s, 3s, 6s, ..., 180s. & \\
            Finally, explain your answer in one sentence. & \\
            \addlinespace
            Here's an example of the response format: & \\
            \addlinespace
            \#\#\# Question: & \\
            How many times did I open the tap? & \\
            \#\#\# Answer: & \\
            You opened the tap twice. & \\
            \#\#\# Evidence: & \\
            $[[9, 15], [120, 135]]$ & \\
            \#\#\# Rationale: & \\
            According to the frames sampled from 9s to 15s, and from 120s to 135s, you opened the tap twice. & \\
            \addlinespace[5pt]
            Your response should strictly follow the example format, including three parts: Answer, Evidence, and Rationale. Do not add any extra content. Here is the question you need to answer: & \\
            \addlinespace[5pt]
            \#\#\#Question:& \\
            \textcolor{teal}{\textless Question\textgreater} & \\
            \addlinespace[5pt]
            \#\#\#Frames: & \\
            The frame sampled at \textcolor{teal}{\textless Second\textgreater}s: & \\
            \textcolor{teal}{\textless Frame\textgreater} & \\
            \textcolor{teal}{$\ldots$} & \\
            \addlinespace
             \VarSty{ {\bf Assistant:} } & \\
        \end{tabular}
    \end{tcolorbox}
    \vspace{-5pt}
    \caption{Prompt for $\mathtt{GPT\,\text{-}\,4o}$ to perform answering and grounding on \smalldatasetname.}
    \label{tab:zs-gpt-4o}
\end{minipage}
\end{table*}

\begin{table*}[!ht]
\centering
\begin{minipage}{0.95\linewidth} % columnwidth
    \vspace{0mm} \centering
    \begin{tcolorbox}%[colback=lightgray!15, colframe=black!50, coltext=black]
        \centering
        \footnotesize
        \begin{tabular}{@{}p{\linewidth} c}
            \VarSty{ {\bf User:} } & \\
            \addlinespace[2pt]
            Frame \textcolor{teal}{\textless Number\textgreater}: \textcolor{teal}{\textless Image\textgreater}. & \\
            $\ldots$ & \\
            \addlinespace
            You are given an ego-centric video. Please watch the video and answer the following question: \textcolor{teal}{\textless Question\textgreater} & \\
            \addlinespace
             \VarSty{ {\bf Assistant:} } & \\
            \textcolor{teal}{\textless Answer\textgreater} & \\
            \addlinespace[5pt]
            \VarSty{ {\bf User:} } & \\
            \addlinespace[2pt]
            The frames are sampled uniformly from 0s to 180s, namely at 6s, 12s, ..., 180s. Localize the time spans that semantically match and support your answer of the question. For example, the answer can be deduced from 10s to 25s, and from 140s to 150s. & \\
            \addlinespace
             \VarSty{ {\bf Assistant:} } & \\
            \textcolor{teal}{\textless Evidence\textgreater} & \\
        \end{tabular}
    \end{tcolorbox}
    \vspace{-5pt}
    \caption{Prompt for InternVL2 to perform answering and grounding on \smalldatasetname.}
    \label{tab:zs-internvl2}
\end{minipage}
\end{table*}

\begin{table*}[!ht]
\centering
\begin{minipage}{0.95\linewidth} % columnwidth
    \vspace{0mm} \centering
    \begin{tcolorbox}%[colback=lightgray!15, colframe=black!50, coltext=black]
        \centering
        \footnotesize
        \begin{tabular}{@{}p{\linewidth} c}
            \VarSty{ {\bf User:} } & \\
            \addlinespace[2pt]
            You are given frames of a video. According to the input frames, answer the following question concisely:  \textcolor{teal}{\textless Question\textgreater}. & \\
            % \addlinespace
            %  \VarSty{ {\bf Assistant:} } & \\
        \end{tabular}
    \end{tcolorbox}
    \vspace{-5pt}
    \caption{Prompt for LLaVa-NeXT-Video to perform only answering on \smalldatasetname. The video inputs are implicitly incorporated with text prompts through function calls.}
    \label{tab:zs-llava-next-video}
\end{minipage}
\end{table*}

\begin{table*}[!ht]
\centering
\begin{minipage}{0.95\linewidth} % columnwidth
    \vspace{0mm} \centering
    \begin{tcolorbox}%[colback=lightgray!15, colframe=black!50, coltext=black]
        \centering
        \footnotesize
        \begin{tabular}{@{}p{\linewidth} c}
            \VarSty{ {\bf System Prompt} } & \\
            \addlinespace[0.1em]
            You are able to understand the visual content that the user provides. Follow the instructions carefully and explain your answers in detail. & \\
            \hrulefill & \\
            \addlinespace
            \VarSty{ {\bf User:} } & \\
            \addlinespace[2pt]
            Frame \textcolor{teal}{\textless Number\textgreater}: \textcolor{teal}{\textless Image\textgreater}. & \\
            $\ldots$ & \\
            \addlinespace
            You are given an egocentric video. Please watch the video and answer the following question: \textcolor{teal}{\textless Question\textgreater} & \\
            \addlinespace
             \VarSty{ {\bf Assistant:} } & \\
            \textcolor{teal}{\textless Answer\textgreater} & \\
            \addlinespace[5pt]
            \VarSty{ {\bf User:} } & \\
            \addlinespace[2pt]
            Detect and report the start and end timestamps of the video segment that semantically matches the textual query \textcolor{teal}{\textless Answer\textgreater} & \\
            \addlinespace
             \VarSty{ {\bf Assistant:} } & \\
            \textcolor{teal}{\textless Evidence\textgreater} & \\
        \end{tabular}
    \end{tcolorbox}
    \vspace{-5pt}
    \caption{Prompt for TimeChat to perform answering and grounding on \smalldatasetname.}
    \label{tab:zs-timechat}
\end{minipage}
\end{table*}

\begin{table*}[!ht]
\centering
\begin{minipage}{0.95\linewidth} % columnwidth
    \vspace{0mm} \centering
    \begin{tcolorbox}%[colback=lightgray!15, colframe=black!50, coltext=black]
        \centering
        \footnotesize
        \begin{tabular}{@{}p{\linewidth} c}
            \VarSty{ {\bf User:} } & \\
            This is a video with 100 frames: \textcolor{teal}{\textless Video\textgreater} & \\
            \addlinespace[2pt]
            \textcolor{teal}{\textless Question\textgreater} & \\
            \addlinespace
             \VarSty{ {\bf Assistant:} } & \\
            \textcolor{teal}{\textless Answer\textgreater} & \\
            \addlinespace[5pt]
            \VarSty{ {\bf User:} } & \\
            \addlinespace[2pt]
            During which frames can we see \textcolor{teal}{\textless Answer\textgreater} happening in the video? & \\
            \addlinespace
             \VarSty{ {\bf Assistant:} } & \\
            \textcolor{teal}{\textless Evidence\textgreater} & \\
        \end{tabular}
    \end{tcolorbox}
    \vspace{-5pt}
    \caption{Prompt for VTimeLLM to perform answering and grounding on \smalldatasetname.}
    \label{tab:zs-vtimellm}
\end{minipage}
\end{table*}

\begin{table*}[!ht]
\centering
\begin{minipage}{0.95\linewidth} % columnwidth
    \vspace{0mm} \centering
    \begin{tcolorbox}%[colback=lightgray!15, colframe=black!50, coltext=black]
        \centering
        \footnotesize
        \begin{tabular}{@{}p{\linewidth} c}
            \VarSty{ {\bf User:} } & \\
            \text{[INST]} \textcolor{teal}{\textless image\textgreater} & \\
            Describe the action in this video frame in one concise sentence. [/INST] & \\
            \addlinespace
             \VarSty{ {\bf Assistant:} } & \\
        \end{tabular}
    \end{tcolorbox}
    \vspace{-5pt}
    \caption{Prompt for the caption module in the multi-stage pipeline to caption the visual content of per-second frames.}
    \label{tab:zs-pipeline-caption}
\end{minipage}
\end{table*}

\begin{table*}[!ht]
\centering
\begin{minipage}{0.95\linewidth} % columnwidth
    \vspace{0mm} \centering
    \begin{tcolorbox}%[colback=lightgray!15, colframe=black!50, coltext=black]
        \centering
        \footnotesize
        \begin{tabular}{@{}p{\linewidth} c}
            \VarSty{ {\bf User:} } & \\
            Here is a video shot from a first-person perspective, recording my activities. I will provide descriptions of each second of the video. Your task is to answer the question I give you based on these video descriptions. Ensure your answer is concise. After answering, provide the time intervals that related to the question-answering as evidence. Evidence is list of intervals (e.g., $[[s1, e1], ..., [s_i, e_i], ...]$) and each time interval (e.g., $[s_i, e_i], s_i <= e_i$) consists of one start timestamp and one end timestamp. Note that evidence may include multiple time intervals. Finally, explain your answer in one sentence. & \\
            \addlinespace
            Here is an example format for your response. & \\
            \#\#\#Question & \\
            When did I open the tap? & \\
            \#\#\#Answer & \\
            You opened the tap from 10s to 20s and 60s to 70s. & \\
            \#\#\#Evidence & \\
            $[[10, 20], [60, 70]]$ & \\
            \#\#\#Rationale & \\
            According to the action descriptions with timestamps, you opened the tap for two times from 10s to 20s and from 60s to 70s. & \\
            \addlinespace
            Here is the video and question you need to review: & \\
            \addlinespace
            \#\#\#Captions & \\
            timestamp, caption & \\
            \textcolor{teal}{\textless timestamp\textgreater, \textless caption\textgreater} & \\
            \textcolor{teal}{$\ldots$} & \\
            \addlinespace[2pt]
            \#\#\#Question & \\
            \textcolor{teal}{\textless Question\textgreater} & \\
            \addlinespace
            Your response should strictly follow the format of given example, including three parts: \#\#\#Answer, \#\#\#Evidence, and \#\#\#Rationale. Do not add any extra content. & \\
            \addlinespace
             \VarSty{ {\bf Assistant:} } & \\
        \end{tabular}
    \end{tcolorbox}
    \vspace{-5pt}
    \caption{Prompt for the reasoning module in the multi-stage pipeline to perform answering and grounding, based on the captions with timestamps acquired from the previous stage.}
    \label{tab:zs-pipeline-reason}
\end{minipage}
\end{table*}

\begin{table*}[!ht]
\centering
\begin{minipage}{0.95\linewidth} % columnwidth
    \vspace{0mm} \centering
    \begin{tcolorbox}%[colback=lightgray!15, colframe=black!50, coltext=black]
        \centering
        \footnotesize
        \begin{tabular}{@{}p{\linewidth} c}
            \VarSty{ {\bf System Prompt} } & \\
            \addlinespace[0.1em]
            You are a helpful assistant designed to output JSON. & \\
            \hrulefill & \\
            \addlinespace
            \VarSty{ {\bf User:} } & \\
            As the instructor of a video understanding course, you have assigned your students to watch a first-person perspective video and answer a question related to its content. Your task is to grade students' answer based on the question and reference answer. Your score should be between 1 and 10, with a higher score indicating a closer match between the student's answer and the reference answer. When grading, consider the following aspects of the student's answer: helpfulness, relevance, accuracy, and level of detail. Provide a brief rationale for your score. & \\
            \addlinespace
            Ensure your response is a dict with `score' and `rationale' as keys. For example, \{`score': 5, `rationale': `\textless rationale \textgreater'\}. & \\
            \addlinespace[8pt]
            \#\#\#Question & \\
            \textcolor{teal}{\textless Question\textgreater} & \\
            \addlinespace[5pt]
            \#\#\#Reference Answer & \\
            \textcolor{teal}{\textless Ground Truth Answer\textgreater} & \\
            \addlinespace[5pt]
            \#\#\#Student Answer & \\
            \textcolor{teal}{\textless Predicted Answer\textgreater} & \\
            \addlinespace
             \VarSty{ {\bf Assistant:} } & \\
        \end{tabular}
    \end{tcolorbox}
    \vspace{-5pt}
    \caption{Prompt for $\mathtt{GPT\,\text{-}\,4o}$ to score open-ended responses based on questions and the associated ground truth answers.}
    \label{tab:gpt4o-evaluate}
\end{minipage}
\end{table*}

\clearpage
\section{Additional Experiments}

\vspace{5pt} \noindent In this section, we present several extended ablation studies and qualitative analysis of the results produced by various models.

\subsection{Extended Ablation Studies}

\paragraph{Ablation of thresholds for generating temporal proposals.}
As outlined in the main paper, we utilize a thresholding method to generate temporal proposals based on both the saliency score vector and the similarity score matrix. 
The threshold value is determined by multiplying a coefficient by the maximum activation value along the time axis. 
In Table~\ref{abalation:thresholds}, we analyze the impact of this coefficient, noting that it varies between the saliency score and the similarity score, because the similarity score undergoes temperature scaling before the application of the softmax function, leading to differing coefficient values.

\paragraph{Training on both egocentric and third-view datasets.}
We train our architecture using both automatically constructed multi-hop triplets and the training split of ActivityNet-RTL. Subsequently, we evaluate the model on \textbf{\smalldatasetname}, as presented in Table~\ref{abalation:mix-multihop}, and on the test split of ActivityNet-RTL, as shown in Table~\ref{abalation:mix-singlehop}.

Our findings suggest that unified training leads to a slight performance decrease across both benchmarks compared to training them separately. 
Despite this decline, the unified model outperforms existing methods on both \smalldatasetname and ActivityNet-RTL. 
This reduction in performance may be attributed to the distribution gap between QA samples and the differing perspectives of the two datasets. 
Incorporating additional grounded QA training data from more diverse sources could potentially enhance the model’s generalization capabilities.

\begin{table}[!ht]
\centering
\begin{minipage}[t]{0.52\linewidth}
\vspace{0pt}
\centering
\begin{subtable}{\linewidth}
\centering
\resizebox{0.86\linewidth}{!}{
\begin{tabular}{lccccc}
\toprule
\specialrule{0em}{2pt}{2pt}
\multirow{2}{*}{\hrulespace\textbf{Strategy}} & \multirow{2}{*}{\hrulespace\textbf{Coefficient}} & \multicolumn{4}{c}{\textbf{Temporal Grounding}} \\ 
\specialrule{0em}{0.3pt}{0.3pt}
\cmidrule(r){3-6}
\specialrule{0em}{0.1pt}{0.2pt}
 &  & mIoP & mIoG & \textbf{IoU@0.3}  & \textbf{mIoU}   \\ 
\midrule
\hrulespace\multirow{3}{*}[-.5em]{Saliency} & 0.6  & 24.0 & 26.7 & 19.1 &  \textbf{15.3}  \\
\addlinespace
& 0.7 & 24.3 & 24.3 & \textbf{19.2} & 14.7  \\
\addlinespace
& 0.8 & 24.8 & 21.9 & 17.3 & 14.0  \\ %[0.15em]
\arrayrulecolor{gray}\hdashline
\specialrule{0em}{2pt}{2pt}
\multirow{3}{*}[-.5em]{Similarity} & 0.05 & 22.2 & 45.0 & 17.4 & 16.3   \\
\addlinespace[1pt]
& \largehrulespace\cellcolor{gray!15}0.10 & \cellcolor{gray!15}23.7 & \cellcolor{gray!15}41.0 & \cellcolor{gray!15}\textbf{18.2} & \cellcolor{gray!15}\textbf{16.7}  \\
\addlinespace
& 0.15 & 24.8 & 32.9 & 16.7 & 15.8  \\
\bottomrule
\end{tabular}
}
\caption{Ablation of thresholds when inferring on \textbf{\smalldatasetname}. 
As detailed in the main text, we determine the threshold for generating temporal proposals by taking the maximum value of the saliency or similarity score for each row vector and multiplying it by a coefficient.}
\label{abalation:thresholds}
\end{subtable}
\end{minipage}%
\hfill
\begin{minipage}[t]{0.43\linewidth}
\vspace{0pt}
\centering
\begin{subtable}{\linewidth}
\centering
\resizebox{0.85\linewidth}{!}{
\begin{tabular}{lccc}
\toprule
\multirow{2}{*}{\hrulespace\textbf{Train Data}} & \multirow{2}{*}{\hrulespace\textbf{\#Frames}} & \multicolumn{2}{c}{\textbf{Temporal Grounding}} \\ 
\specialrule{0em}{0.3pt}{0.3pt}
\cmidrule(r){3-4}
\specialrule{0em}{0.1pt}{0.2pt}
 &  & \textbf{IoU@0.3}  & \textbf{mIoU}   \\ 
\midrule
\hrulespace
Ego & 180 & 18.2 & 16.7 \\
\addlinespace[2pt]
Ego + Third & 180 & 15.3 & 15.1 \\
\bottomrule
\end{tabular}
}
\caption{Effect of training on mixed data for \smalldatasetname.}
\label{abalation:mix-multihop}
\end{subtable}

\vspace{10pt}

\begin{subtable}{\linewidth}
\centering
\resizebox{0.85\linewidth}{!}{
\begin{tabular}{lccc}
\toprule
\multirow{2}{*}{\hrulespace\textbf{Train Data}} & \multirow{2}{*}{\hrulespace\textbf{\#Frames}} & \multicolumn{2}{c}{\textbf{Temporal Grounding}} \\ 
\specialrule{0em}{0.3pt}{0.3pt}
\cmidrule(r){3-4}
\specialrule{0em}{0.1pt}{0.2pt}
 &  & \textbf{mIoU}  & \textbf{P@0.5}   \\ 
\midrule
\hrulespace
Third & 100 & 35.4 & 35.1 \\
\addlinespace[2pt]
Third & 180 & 33.0 & 31.6 \\
\addlinespace[2pt]
Ego + Third & 180 & 32.1 & 27.1 \\
\bottomrule
\end{tabular}
}
\caption{Effect of training on mixed data for ActivityNet-RTL.}
\label{abalation:mix-singlehop}
\end{subtable}
\end{minipage}
\caption{Extended ablation experiments of training and inferring using \modelname.}
\end{table}

\subsection{Qualitative Analysis}

\subsubsection{Visualization of outputs from the saliency branch and similarity branch.}

\begin{figure}[!ht]
    \centering
    \includegraphics[width=.98\linewidth]{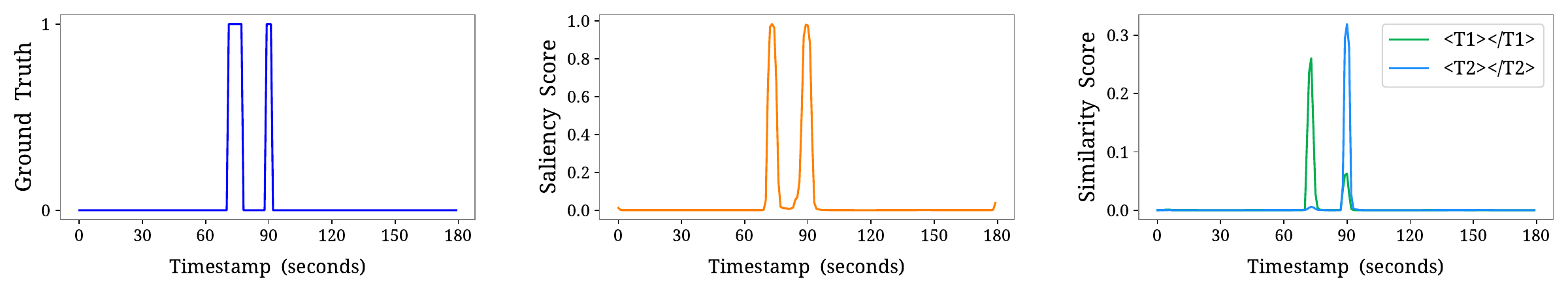} %4bc033f3-9bfe-4b00-8595-09148707bb02_360_540_6
    \vspace{-5pt}
    \caption{Visualization of the ground truth temporal proposals \textit{(left),} along with the results from the saliency branch \textit{(middle)} and similarity branch \textit{(right)} of the \modelname~model to provide temporal evidence for a 2-hop question. 
    }
    \label{fig:branch}
\end{figure}

Given a test video in \smalldatasetname~with the associated query, “What order did I open the fridge and the drawer during the video?”, the response provided by our model is, “You \tstart{1} opened the fridge \tend{1} before \tstart{2} opening the drawer \tend{2}”, 
which involves two distinct time spans. 
As depicted in Figure~\ref{fig:branch}, 
we illustrate the ground truth temporal evidence $\mathcal{T}=\{[71, 78], [89, 92]\}$ in seconds, the saliency score vector $\mathbf{\hat{y}} \in \mathbb{R}^L$ along with each row vector $\mathbf{\hat{S}}_{k, :}$ of the similarity score matrix $\mathbf{\hat{S}} \in \mathbb{R}^{K\times L}$. 
The saliency branch has generated temporal proposals consisting of two windows globally, while the similarity branch is capable of pinpointing the time spans delineated by each pair of grounding tokens within each row of the similarity matrix.

\subsubsection{Visualization of the evaluation results on \textbf{\smalldatasetname}.}

We present additional multi-hop VidQA evaluation results of various models along with our proposed method.

% As illustrated in Figure~\ref{fig:qualitative-result-1}, when posed with the question, ``What tasks did I use the knife for on the chopping board?'', our \modelname~model delivers a semantically similar response with the ground truth answer, 
% ``You cut carrots and peeled off the pumpkin'' and accurately identifies the temporal boundaries for both events as evidence. 
% In contrast, TimeChat provides an incorrect answer and a single, imprecise time interval, and LLaVa-Next-Video offers an approximately correct answer but fails to provide any valid timestamps by replacing different prompts.

% \begin{figure*}[!ht]
%     \centering
%     \includegraphics[width=0.9\linewidth]{figs/results_1.pdf} % ab094ea2-9251-4f10-945b-c2ab00c5282e_180_360
%     % \vspace{-2pt}
%     \caption{The evaluation example about the `Event Composition'.}
%     \label{fig:qualitative-result-1}
%     \vspace{8pt}
% \end{figure*}

\begin{figure*}[!ht]
    \centering
    \includegraphics[width=0.9\linewidth]{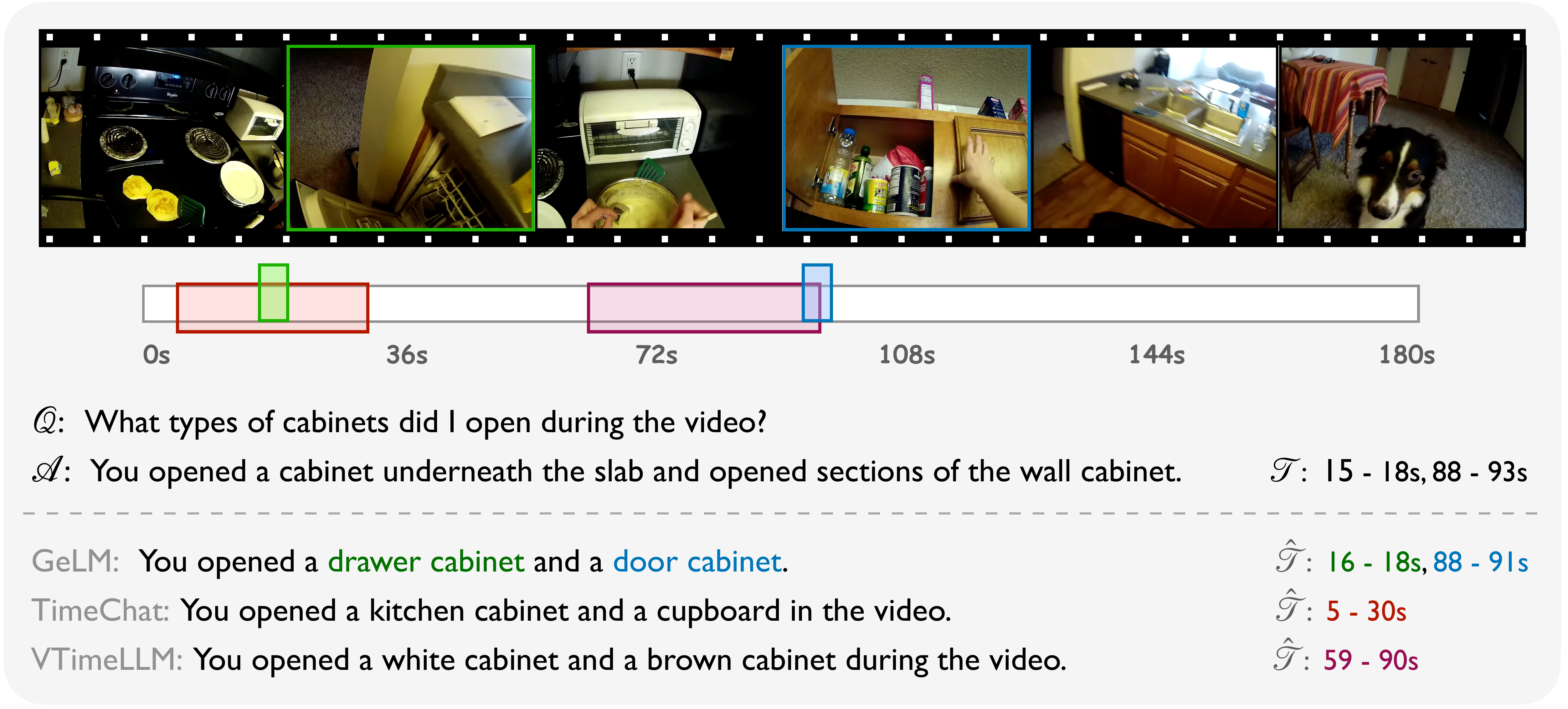} % c707920d-846e-4dc2-9f16-71705507669f_1620_1800
    \caption{The evaluation example about the `Multiple Object'.}
    \label{fig:qualitative-result-1}
    \vspace{8pt}
\end{figure*}

As illustrated in Figure~\ref{fig:qualitative-result-1}, when asked, "What types of cabinets did I open during the video?", our \modelname~model generates the response, "You opened a drawer cabinet and a door cabinet", matching the visual content. Moreover, it precisely identifies the temporal boundaries of both events as visual evidence respectively. In contrast, TimeChat and VTimeLLM offer only a single, imprecise time interval and an ambiguous answer.

\begin{figure*}[!ht]
    \centering
    \includegraphics[width=0.9\linewidth]{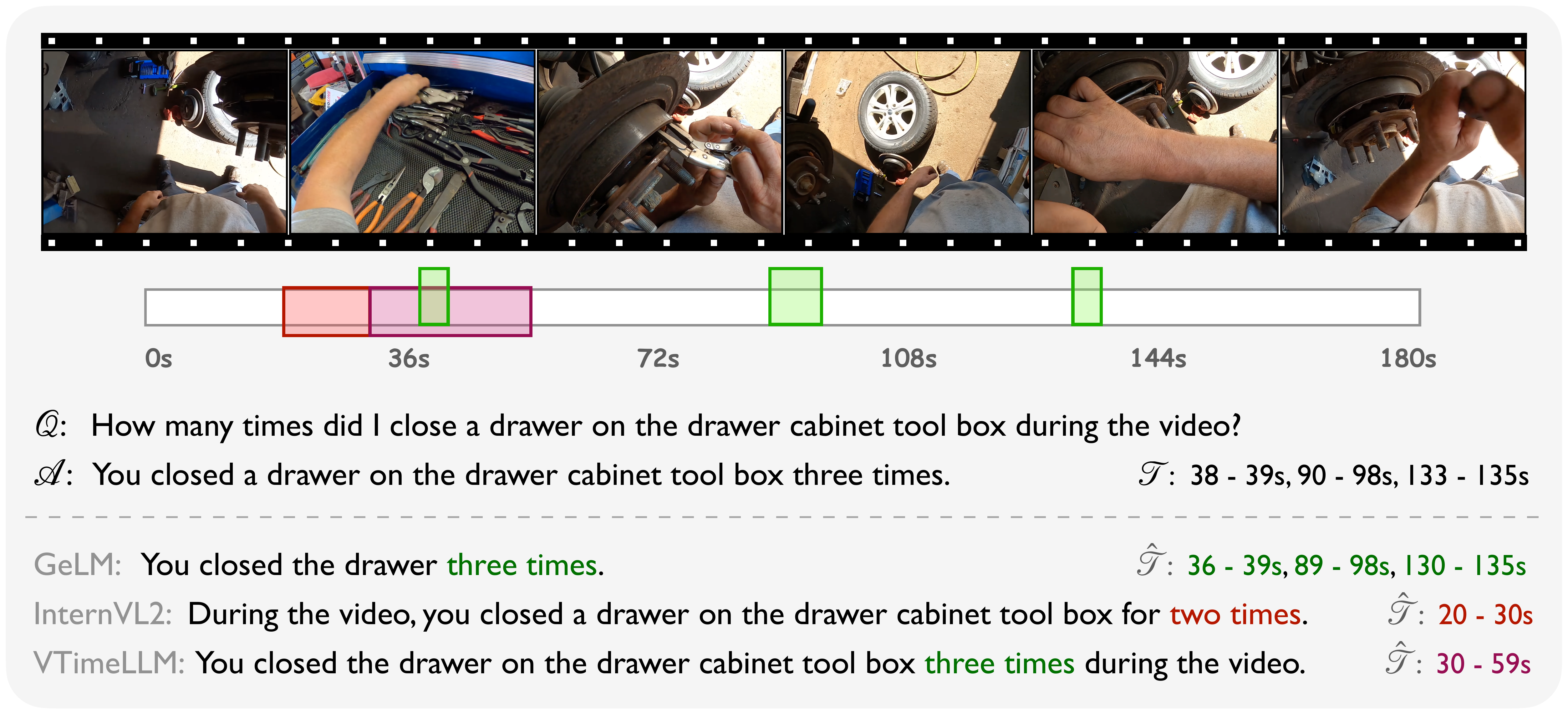} % 252b4e41-56c3-4628-93c9-ab23fea10f09
    % \vspace{-2pt}
    \caption{The evaluation example about the `Repeated Activities'.}
    \label{fig:qualitative-result-2}
\end{figure*}

As shown in Figure~\ref{fig:qualitative-result-2}, 
when asked the question, ``How many times did I close a drawer on the drawer cabinet toolbox during the video?'', our model provides the response ``You closed the drawer three times'', which aligns with the ground truth: ``You closed a drawer on the drawer cabinet toolbox three times''. 
Additionally, it accurately identifies the three specific time spans during which this activity occurs.
However, InternVL2 generates an incorrect answer and time span, even when prompted that multiple time intervals may exist. 
Although VTimeLLM correctly identifies the number of times, it can only provide a single time span, a limitation imposed by its instructions during the fine-tuning stage.

\section{Limitations and Ethical Concerns}

\vspace{5pt} \noindent 
Despite the promising results, 
our proposed method can be improved from three aspects:
Firstly, our automated pipeline has the potential for further enhancement, such as integrating additional visual models to improve the accuracy of data construction. 
Moreover, we suggest extending the application of our automated pipeline to longer videos, and third-person perspective videos, which would better meet the evolving demands of Video LLMs. 
In addition, similar to the approach of calculating mean Average Precision (mAP) separately for small and large objects in object detection, we may consider differentiating the Intersection over Union (IoU) calculations for longer and shorter intervals to more precisely assess the temporal grounding capabilities of various models. 
Finally, incorporating general language grounding data during pre-training may further strengthen the generalization capabilities of our baseline method. With regard to ethical considerations, we recognize that the knowledge embedded in large language models may carry biases related to gender, age, geography, or culture.

% {
% \vspace{30pt}
% \bibliography{aaai25}
% }